Data Visualization to Evaluate and Facilitate Targeted Data Acquisitions in Support of a Real-time Ocean Forecasting System

A Thesis

Submitted to the Graduate Faculty of the
University of New Orleans
in partial fulfillment of the
requirements for the degree of

Master of Science
in
Computer Science

by

Edward Holmberg

August 2014

# Acknowledgements

This research has been supported by the Naval Research Laboratory – Stennis Space Center through Grant N00173-09-1-G032

I would like to express my extreme gratitude to Dr. Germana Peggion who has provided me with so much guidance and patience during the course of this research.

My deepest thanks go to my advisor Dr. Vassil Roussev. Also I would like to thank Dr. N. Adlai A. DePano for his cheerful and always-helpful perspective. Of course, thanks to Dr. Charlie Barron for his continual support and advice.

A shout-out goes to my mother, my family and friends who have provided me with love and moral support throughout this entire process, especially Lily.

Thanks everyone!

# Table of Contents

# List of Figures

# List of Tables

# Abstract

A robust evaluation toolset has been designed for Naval Research Laboratory's Real-Time Ocean Forecasting System RELO with the purpose of facilitating an adaptive sampling strategy and providing a more educated guidance for routing underwater gliders. The major challenges are to integrate into the existing operational system, and provide a bridge between the modeling and operative environments. Visualization is the selected approach and the developed software is divided into 3 packages: The first package is to verify that the glider is actually following the waypoints and to predict the position of the glider for the next cycle's instructions. The second package helps ensure that the delivered waypoints are both useful and feasible. The third package provides the confidence levels for the suggested path. This software's implementation is in Python for portability and modularity to allow easy expansion of new visuals.

# 1. Introduction

The U.S. Navy built and maintains a Real-Time Ocean Forecasting System (RTOFS).  This system predicts future ocean states by using a combination of observations sampled from various instruments such as satellites, buoys, drifters, moorings, and underwater gliders along with a background field.  A data assimilation scheme combines the set of observations and the background field to create an initial state which is the best representation of the ocean at that time.  From this initial ocean state, a dynamical model predicts a future ocean state which can in turn also serve as the next background field for the next set of predictions.

However, uncertainty within the system limits its forecasting accuracy.  An adaptive sampling strategy may be used to improve the forecasting by targeting the next set of observations into areas where the background field yields the most uncertainties.  The sensors used for adaptive sampling are mostly underwater gliders which are programmed with a set of waypoints, or coordinates, that follow a track.

For this research, adaptive sampling is achieved by finding a near-optimal path for the underwater glider by using a Genetic Algorithm (GA) that targets the glider into key areas that would provide a meaningful impact on the system's forecasting capabilities. The GA returns this path as a set of waypoints that may be used by the operative team responsible for deploying the underwater gliders.

Beyond the model-centric criteria that the GA uses, there are additional operative concerns that must also be considered for the glider path such as: avoiding high currents so that it stays on track, avoiding shallow depths, avoiding collisions with other vessels or instruments, and avoiding any water-space exclusion zones.  The current approach for evaluating the GA's optimal solution for these additional operative criteria requires experienced oceanographers to compare the suggested paths to the dynamical model's forecast fields.  Such expertise cannot nor should not be expected of the operative team. Therefore, intelligent guidance of the gliders has to be coordinated between the oceanographers who work with the models and the operative team who make the final decisions for the glider's tour.

Hence, there is a need to implement more qualitative tools that help make those decisions where both the oceanographer and the operative team can use the same tool.  This approach requires visualizations since humans are responsible for performing the evaluations.  The primary goal for this work is to provide robust visualization tools for delivering the glider's next cycle's instructions*.*  The main challenge is to effectively transfer information from the modeling environment to the operative team. The aim is to provide tools that are both simple enough to easily interpret but also complex enough to preserve valuable information to make informed decisions.

The visual evaluation toolset has been subdivided into 3 separate packages because different types of observations may occur at different times and by different types of users during the adaptive sampling operations.

1. **Real-time track versus the suggested path:** The goal is to verify that the gliders are actually following the waypoints and to predict the position of the glider for the next cycle's instructions.
2. **Delivery of useful and feasible waypoints:** The goal is to ensure that the delivered waypoints are both useful and feasible.
3. **An evaluation of the quality of the optimal path:** The goal is to provide the confidence levels for the suggested path.

This paper is organized as follows. In chapter 2 all necessary background information relating to RTOFS and adaptive sampling is detailed. Chapter 3 explains the approach used to visually evaluate the glider suggested path. In Chapter 4, the resulting visualizations are shown along with ways to make them more user-friendly. Chapter 5 explains how to construct a minimal-fit confidence ellipse. Chapter 6 provides an analysis of the effectiveness of the visualizations and future goals.

# 2. Background

This chapter first defines the three fundamental components for an RTOFS and how it creates a forecast. Next, the chapter details adaptive sampling, a method for improving the forecast, and how this method integrates into the RTOFS.

## 2.1   Real-time ocean forecasting system

For a RTOFS, real-time is defined as being capable of using currently available resources to forecast a future state on a daily basis. Such a real-time forecasting system requires 3 fundamental components as illustrated in Figure 2-1: (Robinson, et al, 1998)

1. **Network of observations**, which consist of an assortment of various instruments that collect measurements from the ocean.
2. **Data assimilation scheme**, which adapts this set of observations into an initial ocean model state.
3. **Dynamical model**, which uses the initial state to predict a future state.

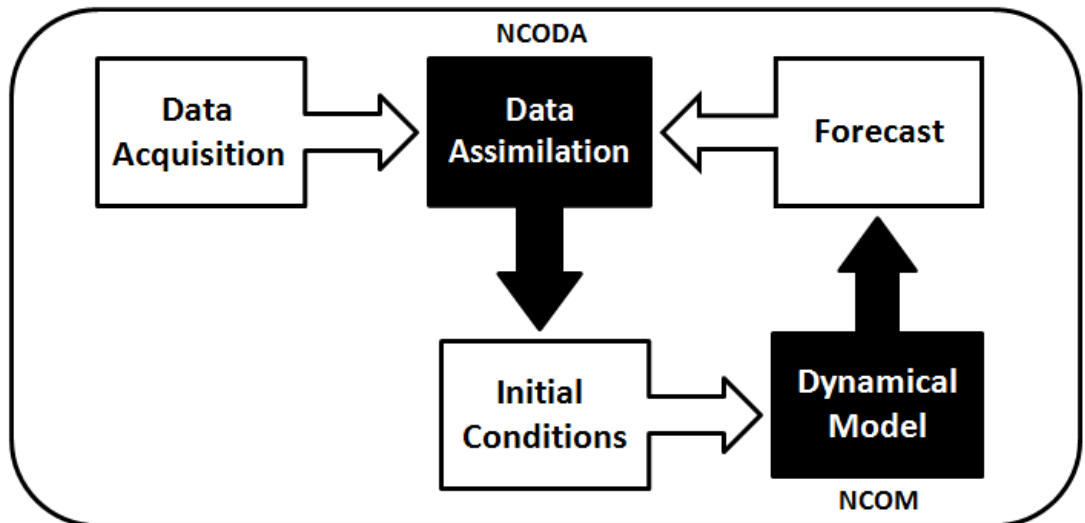

**Figure 2-1:** Real-time Ocean Forecasting System

The observations from the data acquisition phase and a previous forecast, if available, are inputted into the data assimilation model to construct an initial condition. This initial condition is inputted into the dynamical model to create a forecast. This process can then be repeated. The black boxes are models. The white boxes are input/output data.

### 2.1.1   Data acquisition

The first stage to produce a forecast collects data from all the sensors within the network of observations. These instruments can be separated into 3 different categories:

1. **Remote-sensing instruments**, aboard polar-orbiting (global coverage) or geostationary (regional coverage) satellites measure the intensity of electromagnetic radiation at various wavelengths. Specialized combinations of these measurements are used to estimate various properties of the ocean. There are two types of remote observations used in the present system: those that provide sea surface temperatures (SST) and those that provide altimetry, which measures sea surface height (SSH). Data from remote sensing typically covers a broad two-dimensional area of the ocean surface with no information on the subsurface.
2. **In-situ instruments** have high vertical resolution (i.e. depth) but are sparse in 2d space. This type includes expendable bathythermographs (XBTs), conductivity-temperature-depth profilers (CTDs), and underwater gliders. XBTs are small disposable probes that are shot directly into the ocean using a handheld, or ship-mounted, gun-like launcher or dropped from aircraft (AXBT). These probes provide a localized sampling of water temperature and depth, but the standard instruments have a maximum depth of 460 m. An advantage of XBTs is that their deployment does not require the ship to slow down or otherwise interfere with normal operations. In comparison, CTDs are much larger instruments that must be lowered from a ship's deck but they can sample deeper depths and also contain more sensors for collecting salinity, water temperature, pressure, which can be translated into depth, sound speed. Gliders are a type of unmanned, mobile instruments capable of traversing a path in a fully 3d volume. They use Global Positioning System (GPS) to register their surface position for navigation and remotely upload their data, which usually consists of measurements in temperature, salinity, and pressure.
3. **Time series instruments** that provide a high frequency time series of data at a single point in space or along a trajectory. Instruments of this type include moorings, which remain stationary, and drifters, which are free-floating. Drifters typically measure the ocean near the surface; submerged instruments that are passively transported along a density or pressure surface are typically identified as floats. Both moorings and drifters collect ocean data for currents, salinity, and temperature.

### 2.1.2 Data Assimilation

The purpose of data assimilation is to use all the available information to determine as accurately as possible the current state of the ocean, defined on some model grid (Bouttier and Courtier, 1999). To accomplish this, data assimilation balances two sets of inputs: the available observations and a background field. The background field is our best estimate of the state of the ocean prior to the use of the observations. The background information can be generated from the output of a previous analysis or the evolution predicted by a forecast model (Bouttier and Courtier, 1999). While the background completely covers the ocean volume, observations typically sample a small fraction at length and time scales smaller than those the model resolves. When the data assimilation scheme maps the set of observations into the background field, it balances information based on corresponding error or uncertainty estimates which indicate the confidence that these are an accurate representation of the true ocean state at the time and space scales resolved by the forecast system. Mismatches between the observations and background are used to compute innovations to the numerical model, altering the appropriate corresponding model spaces with the new measurements according to estimates of error covariance. The error covariance determines how the model should be adjusted at points where there is no directly corresponding observation; it is a quantity that translates a difference between observations and the background at one point to changes in the model state and confidence over the local area. This scheme must maximize the sparse number of observed spaces so that they make a meaningful impact within the model by having those measurements affect all the approximate neighboring spaces using interpolation and estimation techniques (Bouttier and Courtier, 1999). The data assimilation system used for this research is called Navy Coupled Ocean Data Assimilation (NCODA), developed by J. Cummings. (Cummings, 2005)

### 2.1.3   Forecasting future state

The dynamical model takes this initial state and forecasts a future state. The Navier-Stokes equations are a set of partial differential equations that mathematically describes the dynamics (i.e. the physics) that regulates the ocean evolving state (Peggion, 2007). The ocean is a continuous system, so it must be discretized into a finite number of points; i.e. the numerical model's gridding. We use the numerical values from these "grid" points of the initial state along with the Navier-Stokes equations and any open boundaries or atmospheric forcing to find the new set of values for each of these points. The results from these calculations represent the new future state of ocean within the numerical model. This forecast can then be used as the next background field for the next set of data assimilations. The dynamical model used for this research project is called Navy Coastal Ocean Model (NCOM), developed by P. Martin. (Martin, 2000; Barron et al., 2006)

## 2.2   Improving the forecasting

Using the update cycle, the ocean forecasting system predicts future states of the ocean. However, the accuracy of these predictions is hindered due to limitations that exist within the system. The limiting factor for forecasting are areas within the system that contain uncertainties which result in a deviation of the model's forecast from the actual ocean state (Thunnissen, 2003). *'All oceanic dynamical models are imperfect, with errors arising from: the approximate physics which govern the explicit evolution of the state variables, the approximate physics which parameterizes the interaction of the state variables and the discretization of continuum dynamics into a numerical model'* (Robinson, et al., 1998, p.544). Therefore, the next objective is to improve the system's forecasting capabilities, as illustrated in Figure 2-2. The goal is to find a way to minimize the uncertainties affect on the system. However, the advantage of the data assimilation scheme is that the relative uncertainties of the dynamics can be corrected with the integration of the next set of observations (Robinson et al, 1998). Thus, improved forecasting can be accomplished by improving the quality of the data collected from the set of observations that are used during the data assimilation phase to construct the initial state. For this purpose, quality data may be defined as those observations that provide the most impact on the system's ability to forecast. It is necessary to target collections of data that can give useful information for updating the areas of uncertainty. This practice of targeted sampling to correct the forecasting system is called adaptive sampling (Leonard, et al, 2003). Adaptive sampling requires that the areas of interest be identified so that they can be targeted for the next set of observations. In this research, this is accomplished using a GA based on a cost function that is defined upon the uncertainties from the system's previous forecast. This cost function is then used to identify areas of interest. A GA will return a set of waypoints, or coordinates, that maximize this cost function and are therefore optimal by that definition. The next data acquisition will then direct sensors to sample those areas of highest interest. (Smedstad et al, 2012)

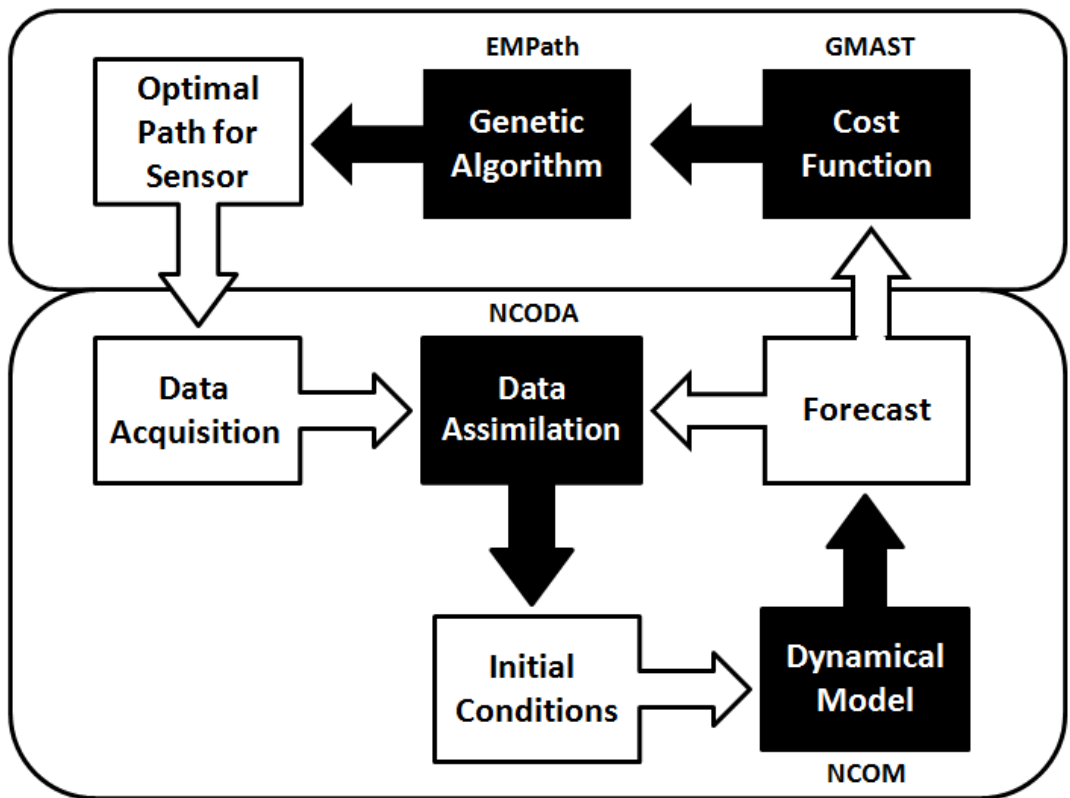

**Figure 2-2:** Improved Real-time Forecasting System
The forecast from Figure 2-1 is used to construct a cost function. That cost function is used by the GA to find the best use for the sensors for the next set of data acquisitions. The black boxes are models. The white boxes are input/output data.

## 2.2.1 Constructing a cost function

The purpose of the cost function is to identify areas in the system where more data acquisitions will have the largest impact on improving the forecast. This implies that the relative value of different observations must be quantified. To do this, the contributing factors that cause doubt within the forecast must be identified and used to define the set of constituent cost functions (CCF). A CCF defines an attribute within the forecast that affects its accuracy. It is ideal to identify every possible, doubt-inducing attribute as its own individual CCF. The CCFs are based upon three criteria:

1. ***Model forecast uncertainty*** – the more effective observations are those cases wherein a small change in a value of the numerical model leads to a big difference in the resulting forecast. (Smedstad et al., 2012)
2. ***Ocean temporal-spatial variability*** – the more effective observations are those that better quantify and reduce uncertainty in areas that change a lot due to the evolution of the features. This variability can be divided into two types: (Heaney, 2010)
    a. *Temporal Variability*, which are the changes in an area over time caused by the dynamics.
    b. *Spatial Variability,* which are the changes that occur across the space caused by gradients, fronts, or eddies.
3. ***Operation constraints*** are physical, political, or logistical reasons why possible sampling-spaces or configurations should be avoided. These may include ensuring that the distance between multiple sensors is maintained to avoid collisions or oversampling from the same location (Heaney, 2007) avoiding shallow depths or strong currents and geographic boundaries, such as international borders or water-space exclusion zones associated with other naval or shipping activities. (Heaney, 2013)

Finally, all these CCFs are then linearly combined to form a global cost function; wherein each individual CCF is assigned some weight. This global cost function is used to identify the 'best' sampling areas. This is crucial because the solutions of the GA optimization are only as good as the cost function that is defining them. (Heaney, 2007)

**2.2.2 Genetic algorithm**

In terms of computational complexity, calculating a sensor's optimal path, one that collects the maximal amount of quality data, is considered an *NP-hard problem*; this means that as the problem size increases the solution time increases exponentially (Popaf et al., 2004). For such a problem where an exact answer is too difficult to be analytically calculated an optimization technique called GA can be used to find a best, near-optimal path (Heaney, 2013). The basic premise for the GA uses the concepts of survival-of-the-fittest whereby a number of different possible paths are randomly generated and then ranked. Those highest-ranked paths are kept and the rest are discarded. Afterwards, those paths that were kept are then mutated and reevaluated along with some newly generated random paths. This process continues across many generations until the program terminates and the highest-ranked path approximates the optimal path to route the sensor. The path rankings are done using the cost function constructed from the dynamical model's forecast (Smedstad et al., 2012). The GA used for this research project is called Environmental Measurements Path Planner, (EMPath), developed by K. Heaney. (Heaney et al., 2012)

### 2.2.3 Underwater Gliders

Underwater gliders are ideal instruments, which meet the criteria to perform many tasks necessary for adaptive sampling operations. They are mobile instruments that may maneuver from one location to another while tracked with GPS and may be equipped with a variety of sensors capable of acquiring whatever type of data is desired. Underwater gliders are energy efficient and designed for continuous, long-term deployment. They are able to provide sustained observations over vast ocean regions. In addition, underwater gliders allow high horizontal and vertical sampling resolutions, making them able to detect small-scale features along their assigned path (Mourre and Alvarez, 2012). Since underwater gliders are unmanned, they must be programmed, before being deployed, but this allows them to operate autonomously without requiring constant human oversight. Glider missions are typically planned to reach a series of locations commonly called waypoints. The possibility of freely selecting the mission waypoints, so that the data are collected at optimal locations to maximize their information content, has led to the concept of glider adaptive sampling (Mourre and Alvarez, 2012). Additionally, since gliders are inexpensive, they can be deployed in large numbers called a glider fleet. The advantage of deploying a glider fleet is that they can work in unison to maximize the sampling of quality data.

However, despite all these advantages, there are two main constraints that may restrict the glider operations. The first main constraint is that gliders have low-powered motors and limited battery resources. They must avoid traversing against high currents, otherwise they may deviate off course or miss the desired waypoints (Mourre and Alvarez, 2012). Compounding this problem, gliders only communicate while on the surface. This communication constraint leads to intermittent feedback, which renders the task of coordination challenging. Since the position and estimated gradient information are not available continuously this necessitates the need for a method that detects if a deviation from schedule has occurred and if so then how best to correct it (Fiorelli, et al, 2003). The second main constraint is the time available for the glider mission before its rendezvous' point where it is recollected. Additionally, during an exercise, the glider is assigned an area of operation where it is required to stay. Inside this area there might be additional constraints such as currents, shallow water, and bathymetry that all must be taken into account while preparing the input parameters for running EMPath.

## 2.3  Problem statement

The oceanographers, who are model-oriented, are responsible for taking care of the quality of the solutions. But it is the operative team who provides the final instructions for the glider's next sampling cycle. The oceanographers must communicate the next set of inputs to the naval pilots, who are responsible for ensuring that the coordinates programmed into the sensors are feasible. For this purpose, the aim is to provide a visual evaluation tool set from which both users may benefit.

**2.3.1 Requirements**

This visual evaluation toolset must be capable of providing the necessary insights to the operative team in the form of images, animations, and text. To ensure consistency it must interface with the existing forecasting system to produce graphics. The toolset must be portable so that it can run across multiple platforms. It must be intuitive and easy to use without confusing the end user. It must also offer options to customize the output to make it user friendly.

# 3. Approach

Chapter 3 explains the approach for developing the three visualization packages. Section 3.1 describes the Naval Research Laboratory's RTOFS which is called RELO and the datasets that are the input for the visualization packages. Section 3.2 explains the modular programming approach used to implement the visual evaluation toolset where individual software modules are written to access and use each of these data components from 3.1. In section 3.3, the three visual evaluation packages' executables are detailed along with the user options. The output files from this evaluation toolsets are formatted such that: images are delivered as png files, tables are delivered as text files, and animations are delivered as gif files.

Python is the programming language of choice because it is suitable for scientific computing and meets most other requirements:

1. The software must utilize the pre-existing data, configurations and frameworks of the operational systems. It must support I/O operations for multiple file formats such as: netCDF, text, bin, csv and png.
2. The software must handle batchmode-style executions and perform OS-level 'file directory' tasks which include traversing, creating, and removing files and directories.
3. The software must be portable, where it is platform agnostic.

## 3.1 Input data from the RELO RTOFS

This section provides a summary of the operational systems with a focus on the data utilized for the visualization packages.

### 3.1.1 Relocatable Circulation Prediction System (RELO)

The NRL operational ocean forecasting system is RELO. RELO has two major components (Coelho et al., 2009):

1. *Navy Coupled Ocean Data Assimilation (NCODA)* which is used for the data analysis and model initialization (as described in section 2.1.2).
2. *Navy Coastal Ocean Model (NCOM)* which is used for the ocean dynamics predictions (as described in section 2.1.3).

RELO is a vast and complex system with numerous outputs in several file formats. The most relevant for this research are:

1. *Forecasted Fields* - The netCDF files contain the predicted fields at given time increments.
2. *Bathy file* - RELO provides an auxiliary program which produces a static bathymetry matrix variable in netCDF file format using the water depth values

### 3.1.2 Glider Mission Adaptation Strategies (GMAST)

GMAST translates a glider sampling strategy into criteria for evaluating alternative glider paths through EMPath (Smedstad, et al, 2012). These criteria manifest as a set of CCFs that are derived from the RELO netCDF's forecast fields. GMAST's only function is to produce the CCF file henceforth referred to as GMAST_CCF

1. *CCF file*
   The GMAST_CCF is a netCDF file containing the CCFs which aims to highlight the model uncertainty and ocean variability, as detailed in section 2.2.1. The longitude and latitude spacing uses RELO's model gridding. This netCDF file has 3 major components:

   i. *CCFs* - There are two different types of CCFs: static and dynamic. Static cost functions have no time dimension whereas dynamic cost functions do. To identify ocean variability the CCFs are based on the mean and standard deviations from predicted salinity and temperature fields. Other constraint-based CCFs, such as a rendezvous point, may also be included.
   ii. *Water currents* - Water currents are dynamic functions that are averaged over the glider depth of the NS-EW velocity component. These values are in terms of meters per second.
   iii. *Metadata* – These are additional parameters relating to the netCDF file and the glider mission information. Among those parameters the most prominent for this project are:
       a. *Start_time* – the assumed time that new instructions are given to the glider
       b. *DeltaTime* – the time increment of the dynamic functions.

### 3.1.3 Environmental Measurements Path Planner (EMPath)

EMPath is the software package that implements the GA used for coordinating adaptive sampling (Heaney et al., 2012). The goal is to provide a visualization package to better evaluate and illustrate the EMPath results. EMPath is the last system in the chain before constructing the visualizations. EMPath implements the GA in the following way: A number of individual possible paths are randomly created where the population size is a parameter called *individuals*. The GA then iterates over this population, improving the paths using the techniques from 2.2.2. The number of iterations that are performed is based on a parameter called *generations* wherein the top path is the final result of the GA. EMPath runs this GA sequence for multiple times, where the number of times the GA is performed is based on a parameter called *runs*. Each run produces its own top path and the path with the highest score from them all is called the Best run.

EMPath is a self-contained program that runs independent of either RELO or GMAST thereby requiring that the necessary data must be passed in as parameters. EMPath only requires 4 major inputs:

1. *GMAST_CCF* from 3.1.2
2. *Bathy_file* from 3.1.1
3. *Input.prm file* which contains all the necessary parameters for the EMPath executions
4. *Cords_init file* which provides the initial starting position of the glider.

EMPath produces 4 major outputs:

1. *Morphology.txt* - This text file contains the morphology values for each of the latitude and longitude coordinates at the initial time, (time index=0). This file contains each of the CCFs and the final column is the combined cost function. (Heaney et al., 2012)
2. *GA_Run#.csv* - For each of the runs, there is a csv file that contains all of the details of the top path. These details include the lon, lat, and bearing for every glider at every time. (Heaney et al., 2012)
3. *GA_BestRun.csv* - GA_BestRun.csv is the GA_Run#.csv that has the highest score. (Heaney et al., 2012)
4. *EMPath.log* - EMPath's standard out is piped to a text file. This log file contains the scores for all the runs.

## 3.2 Modular design

The visualization packages are designed using a modular programming approach which is a technique that separates the functionality of a program into independent, interchangeable modules. Each package contains everything necessary to execute only one aspect of the desired functionality. Modules can be separated into three types:

1. *Base modules* serve to provide accessibility and functionality for the input data into the visual evaluation toolset. Each of the data inputs listed from 3.1 has a base module implemented just for it.
2. C*omposite modules* are made from a combination of the *Base modules* to produce a visualization used for the evaluations.
3. *Main module* is the executable for each of the packages that provides all of the visualizations that must be delivered

The advantage of a modular approach is that new visualizations may easily be added and that existing modules may easily be upgraded. Such a design also allows for quick prototyping while also establishing a flexible, uncoupled system.

### 3.2.1 Modular hierarchies for visualization packages

#### A  Run_tracker

The user executes the run_tracker script which creates the visualizations described in 4.1.  Figure 3-1, shows the modular hierarchy for the first package's main module called *run_tracker*. *Run_tracker* uses 2 composite modules called *Plot Track vs. Path* and *Predict Glider*.  These composite modules are responsible for building the necessary graphics for the user. These composite modules require the base modules to produce its output.

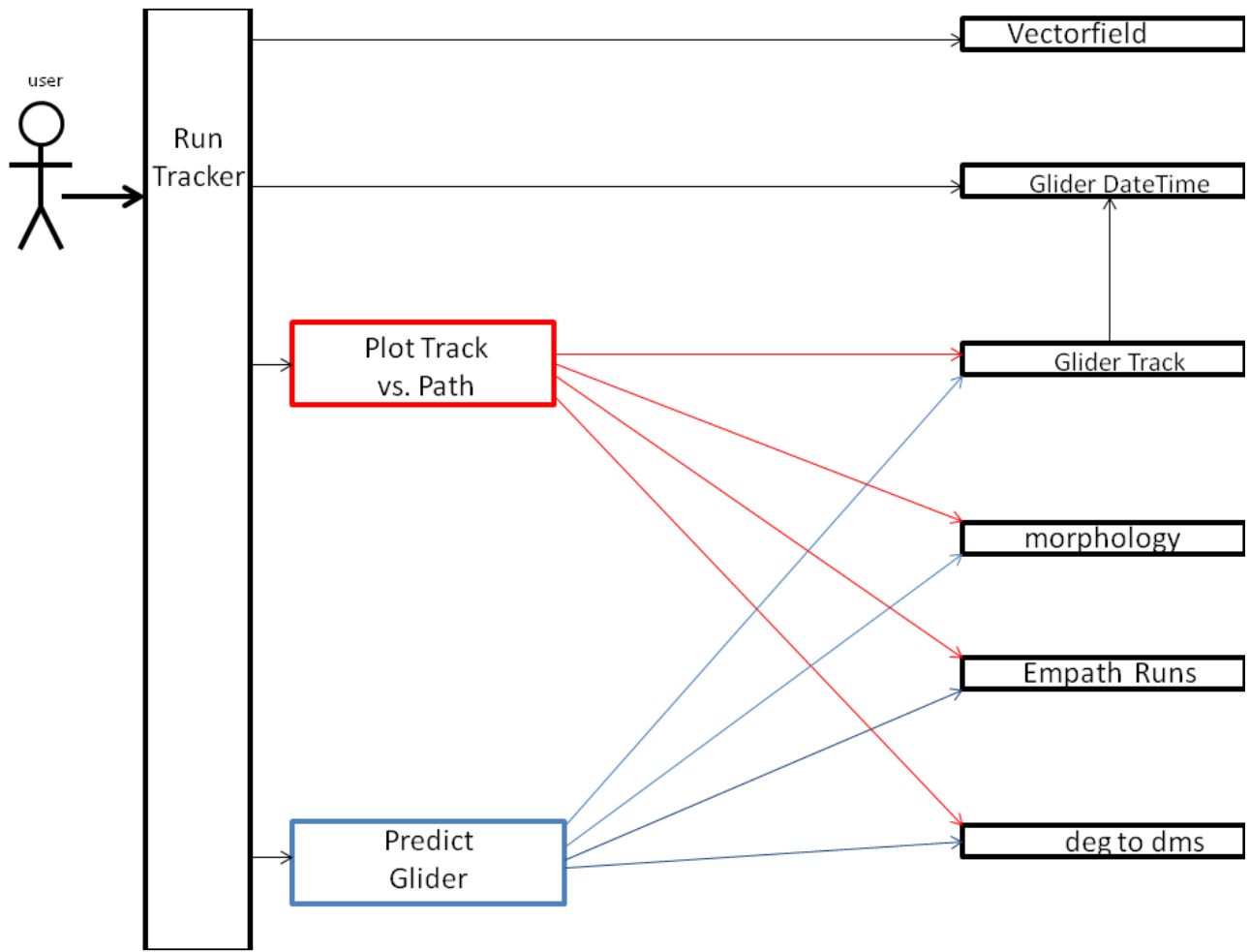

**Figure 3-1**  The interactions diagram for run_tracker and the various modules it uses.  The composite modules are colored and the base modules are black. The executable is also black. The colored arrows show the dependencies for the corresponding composite module

## B Run_visuals

The user executes the run_visuals script which creates the visualizations described in 4.2. Figure 3-2, shows the modular hierarchy for the executable script which calls 4 external composite modules that are used for building each of the individual visualizations for this package. The composite modules are dependent on the base modules to construct each of the images.

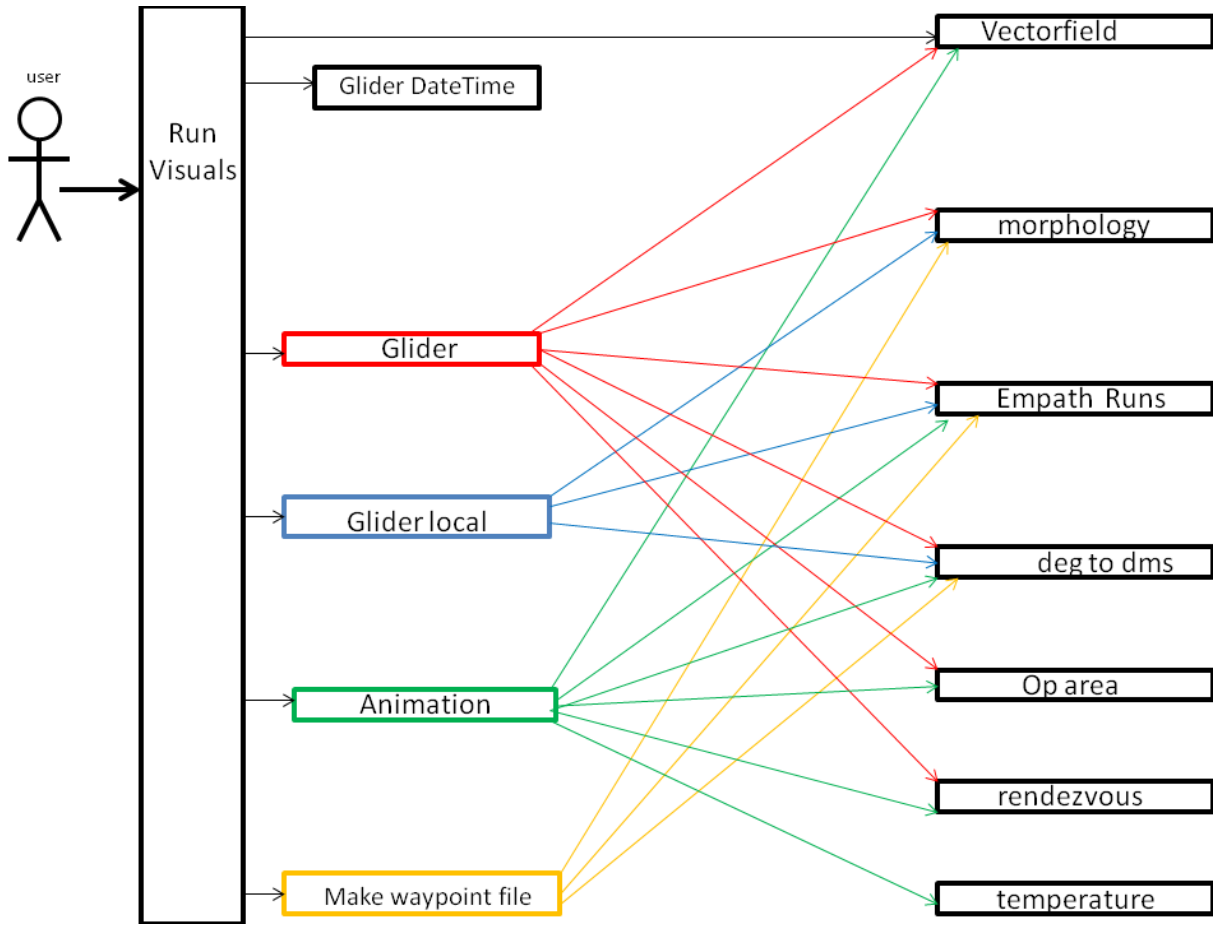

**Figure 3-2** The interactions diagram for run_visuals and the various modules it uses. The composite modules are colored and the base modules are black. The executable is also black. The colored arrows show the dependencies for the corresponding composite module

## C Run_evaluations

The user executes the run_visuals script which creates the visualizations described in 4.3. Figure 3-3, shows the modular hierarchy for the executable script which calls 5 external composite modules that are used for building each of the individual visualizations for this package. The composite modules are dependent on the base modules to construct each of the images.

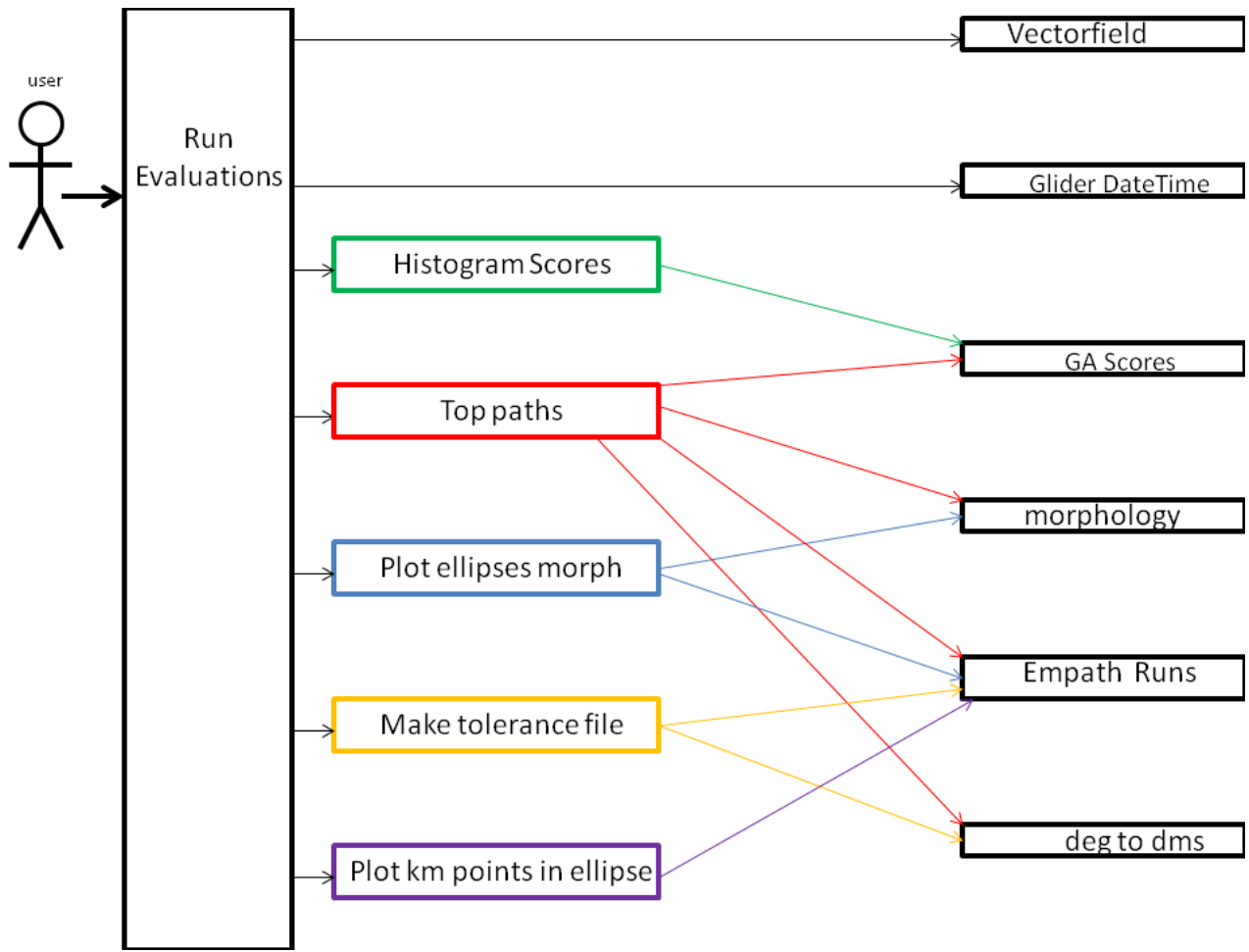

**Figure 3-3** The interactions diagram for run_evaluations and the various modules it uses. The composite modules are colored and the base modules are black. The executable is also black. The colored arrows show the dependencies for the corresponding composite module

## 3.3 Executing the visualizations

The main modules are python scripts that contain the user parameters and calls to the composited modules. The parameters are passed from the main module to the composite modules. The software is designed to have an easy interface for user input. All the common parameters are contained in a file to provide consistent access to the data. This file is called *common_params.* Table 3-1, shows the parameters contained within this file.

```
#Path to all default Python files
BINDEF = '/net/vulcan/export/scratch2/holmberg/bin.v2/'

#####################[ DIRECTORY Params ]##################################
Glider_names = ['Alfa', 'Bravo', 'Clint', 'Delta', 'Echo', 'Fox', 'Golf', 'Hotel']

ensrun = 0                                   ##Declare run 'type'     | range: (0=Control,
1=Ensemble)
REGION = 'nest1'                             ##Declare region name    | range: string
```
**Table 3-1**: common parameters that all 3 visualization packages use

### 3.3.1 run_tracker

This first set of data visualizations are produced by executing the python script: *run_tracker.* As shown in Figure 3-1, this script calls auxiliary composite modules to produce the visuals. Table 3-2 illustrates the RELO data that this package requires. The options for this visualization tool are shown in Table 3-3. Table 3-2 summarizes the files that are used for the 1st visual tool (4.1).

| Input file | File type | Data description |
|---|---|---|
| **Glider log files** | text | records of the glider's coordinates with a timestamp |
| **EMPath's 'best run' data** | csv | provides the suggested optimal path for the glider |
| **EMPath's morphology** | text | contains the cost function values that highlight the areas of uncertainty |
| **EMPath's input parameters** | text | provides the estimated gliders' speed in meters per second |

**Table 3-2**: The table displays the source input file, the file format and the description for that data

```
'''This file supplies all user-defined parameters to the Empath 'Data Tracker' scripts '''
#####################[ Image Params ]##################################

#Options for Axes Labels
set_dms = 1       ##Sets x-axis & y-axis to use deg,min,sec   | range: (0=FALSE, 1=TRUE)

#Options for Image contents
Show_Gliders   = 1    ##Shows Glider paths,                   | range: (0=FALSE, 1=TRUE)

#Options for Forecasting
Max_track_hours = 148      ## Use X Hours of glider's real track | range: (0<X)
Max_path_hours  = 48       ## Use Y Hours of EMPath waypoints    | range: (0<Y)

#Options for Morphology
Smooth_Image   = 1    ##Smooths pixels of the morph image,  | range: (0,1,2,3,4) where 0=none, 4=most
```
**Table 3-3**: The user parameters for the first visualization package

### 3.3.2 run_visualization

The second set of data visualizations are produced by executing the python script: ***run_visualizations.py.*** As shown in Figure 3-2, this script calls auxiliary composite modules to produce the visuals. Table 3-4 illustrates the RELO data that this package requires. The options for this visualization tool are shown in Table 3-5. Table 3-4 summarizes the files that are used for the 2nd visual tool (4.2).

| Input file | File type | Data description |
|---|---|---|
| **GMAST CCF** | netCDF | contains the 'water currents' forecasts which can be used to determine those areas with strong currents to avoid |
| **EMPath's 'best run' data** | csv | provides the suggested optimal path for the glider |
| **EMPath's morphology** | text | contains the cost function values that highlight the areas of uncertainty |
| **EMPath's input parameters** | text | provides the operational area which dictates and restricts where glider can go |

**Table 3-4:** The table displays the source input file, the file format and the description for that data

```
'''This file supplies all user-defined parameters to the EMPath 'Data Visualization' scripts '''
########################[ Image Params ]################################

#Options for Axes Labels
set_dms = 1         ##Sets x-axis & y-axis to use deg,min,sec  | range: (0=FALSE, 1=TRUE)

#Options for Image contents
Show_Vectors   = 1      ##Shows Vector fields,         | range: (0=FALSE, 1=TRUE)
Show_Gliders   = 1      ##Shows Glider paths,          | range: (0=FALSE, 1=TRUE)
Show_OpArea    = 1      ##Shows Operational Area,      | range: (0=FALSE, 1=TRUE)
Show_SingleGL  = 1      ##Make single glider images    | range: (0=FALSE, 1=TRUE)

#Options for Vector images
Field_Density = 3       ##The gap size between vectors,  | range: (1,2,3,4,...)
Mask_Vectors  = 0.      ##Masks vectors under this size  | range: (x >= 0; in m/s)
Ref_Vector    = 1       ##Set the reference vector size  | range: (x > 0; in m/s)

#Options for Morphology
Crop_Morph     = 3   ##Sets 'n' outer LON/LAT vals as NaN, | range: (0,1,2,...)
Show_Colorbar  = 0   ##Shows the color bar,                | range: (0=FALSE, 1=TRUE)
Smooth_Image   = 1   ##Smooths pixels of the morph image,  | range: (0,1,2,3,4) where 0=none, 4=most

########################[ Animation Params ]################################
Make_Animation = 1 ##Make glider animation with SST | range: (0=FALSE, 1=TRUE)
Ani_Ref_Vector = 1 ##Set the reference vector size  | range: (x > 0; in m/s)
```

**Table 3-5**: The user parameters for the second visualization package

### 3.3.3 run_evaluations

The third set of data visualizations are produced by executing the python script: *run_evaluations.py.* As shown in Figure 3-3, this script calls auxiliary composite modules to produce the visuals. Table 3-6 illustrates the RELO data that this package requires. The options for this visualization tool are shown in Table 3-7. Table 3-6 summarizes the files that are used for the 3rd visual tool (4.3).

| Input file | File type | Data description |
|---|---|---|
| **EMPath's log file** | text | contains the scores for each run |
| **EMPath's run data** | csv | includes all of the suggested paths for the gliders |
| **EMPath's morphology** | text | contains the cost function values that highlight the areas of uncertainty |
| **EMPath's input parameters** | text | provides the estimated gliders' speed in meters per second |

**Table 3-6:** The table displays the source input file, the file format and the description for that data

```
'''This file supplies all user-defined parameters to the 'Path Evaluations' scripts '''
#######################[ Image Params ]##################################

#Options for Axes Labels
set_dms = 1      ##Sets x-axis & y-axis to use deg,min,sec   | range: (0=FALSE, 1=TRUE)

#Options for which waypoints to evaluate
Start_hour = 12     ## Hour of 1st WP to evaluate            | range: (0<X)
Stop_hour  = 48     ## Hour of last WP to evaluate           | range: (X<=Y)
STD        = 1.5    ## STD steps to include in ellipse       | range: (0<STD<3)

#Options for Morphology
Smooth_Image  = 1   ##Smooths pixels of the morph image,     | range: (0,1,2,3,4) where 0=none, 4=most
```

**Table 3-7**: The user parameters for the third visualization package

# 4. Methods with results

The visual evaluation toolset has been subdivided into 3 separate packages because different types of observations may occur at different times and by different types of users during the adaptive sampling process. There are 3 different tasks that must be performed:

1. Real-time track versus the suggested path
2. Delivery of useful and feasible waypoints
3. An evaluation of the quality of the optimal path

## 4.1 Map-based comparison of real-time glider track vs. suggested path

The objectives are i) to focus on monitoring and managing the gliders after their deployment and to ensure that they move into position to collect meaningful data, and ii) that they remain on schedule according to their projected tours. The following paragraphs describe the 2 major associated visualizations.

### A. Real Track versus suggested path
The goal is to show that the glider is following the intended path as instructed within the expected time frames.

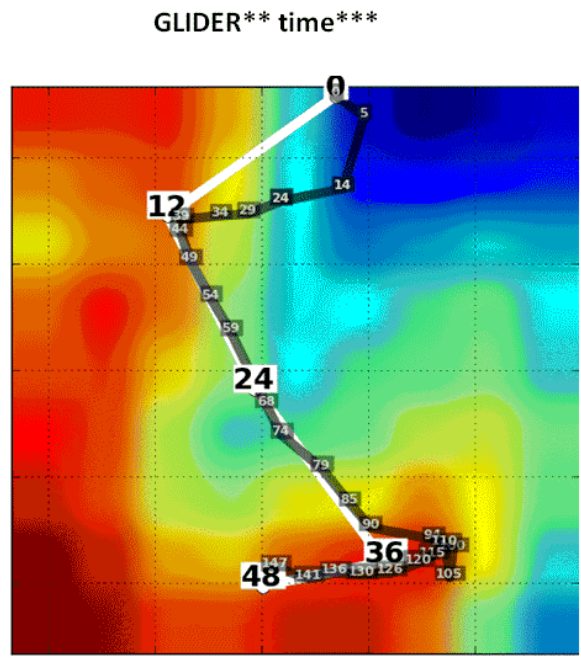

**Figure 4-1:** The actual glider track with report-in times (semi-transparent black); the waypoints of the suggested path (white) and morphology (background colors.) The coordinates and time have been removed and are not available for public release.

This verification is accomplished by comparing the gliders' tracks to the waypoints of EMPath's suggested path. If the glider is off the path, then corrections may be made for the next set of instructions. EMPath's suggested path is plotted using its waypoints as illustrated in Figure 4-1. The waypoints are labeled with each expected hour-of-arrival from the initial time of deployment. The glider's actual track is then constructed using the log files that it regularly uploads whenever it surfaces and records its positions along with a timestamp.

### B. Predicted position for next set of instructions

The goal is to predict the position of the glider at the next cycle's instructions. This predicted position may be used as the initial coordinates for the next run of EMPath. This allows generating the next cycle's instructions from a more accurate expectation of the gilder's position. As shown in Figure 4-2, the glider's last reported position is used along with the to-be-reached waypoints to predict where it's likely to be for the next set of instructions. This is calculated by taking the glider's current coordinates and the next waypoints' coordinates and finding the distances between them in kilometers. By using the glider's speed the glider's future position can be predicted assuming the glider moves at a constant rate and the ocean is at rest. This way if the glider's track is different from the suggested path, then a correction is made to reflect this for the next set of instructions.

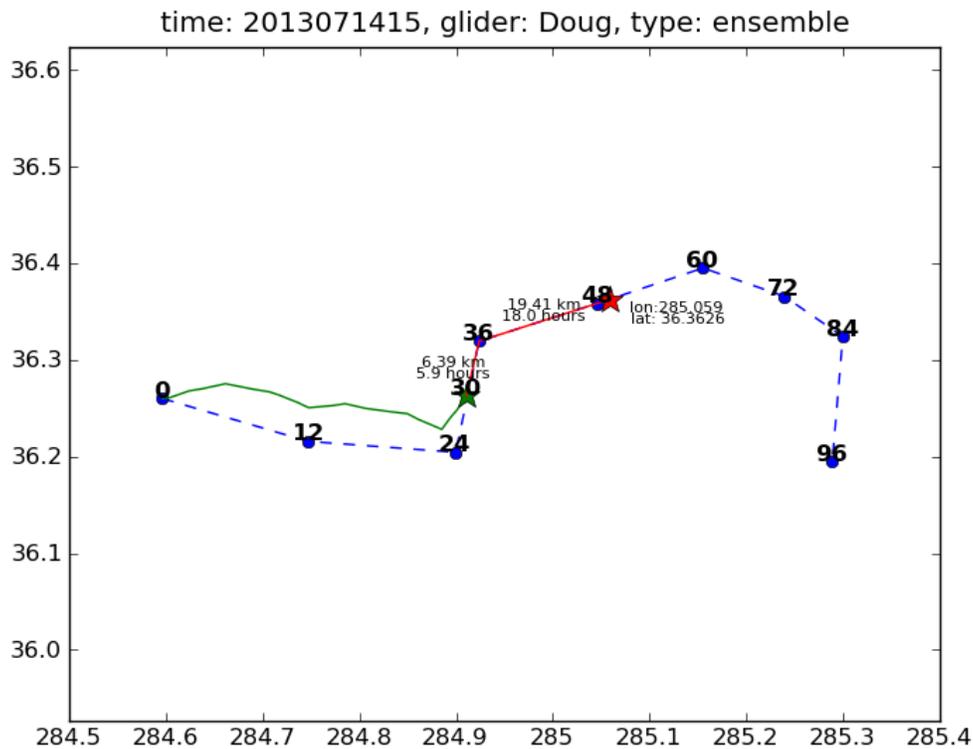

**Figure 4-2:** Comparison of the suggested path (dashed line) and the predicted path (red line) from the last recorded position (green star)

### 4.1.1 Options

Options are included to make the plots more user-friendly. All 3 visualization packages options are editable by the user and located in the respective *run* script.

1. *Max Track Hours* - By default, it draws all of the available track data. If the glider has too much tracking data, then it can be difficult to see the relevant segment of its track. This option only shows the relevant segment to draw.

2. *Max Path Hours* - From the delivered waypoints, this option only uses those waypoints up to the given hour. This way only the waypoints relevant to the glider's current progress are displayed for the comparison.

3. *Show Glider Path* - This option toggles the ability to plot the EMPath suggested path on or off. The glider's track is always plotted.

## 4.2: The delivery of useful and feasible waypoints.

The goal is to determine if the waypoints are useful and feasible. 'Useful' because they should sample the areas that have the most impact on the model. 'Feasible' because the sampling sensors must be able to reasonably follow the suggested path. If the waypoints don't meet these conditions EMPath may be rerun with different input criteria to produce a new set of waypoints before the final delivery to the operative unit. The following paragraphs describe the four major associated visualizations.

### A. Full Morphology with the Operational Area and All Glider Paths

The goal is to depict the path versus the morphology possibly including oceanographic limitations such as the bathymetry and currents. The morphology's values are normalized and color mapped. As illustrated in Figure 4-3, high values (red) of the morphology are associated with areas of uncertainty. Assessing the feasibility of the path considers the gliders' limitations as discussed in section 1.2.3. The operational area and velocities depicted in Figure 4-3 help determine if the path is 'feasible.' For example if the suggested path interacts with strong currents or shallow waters then the operational area may be readjusted to exclude those areas and resubmitted to EMPath to generate a new path plan.

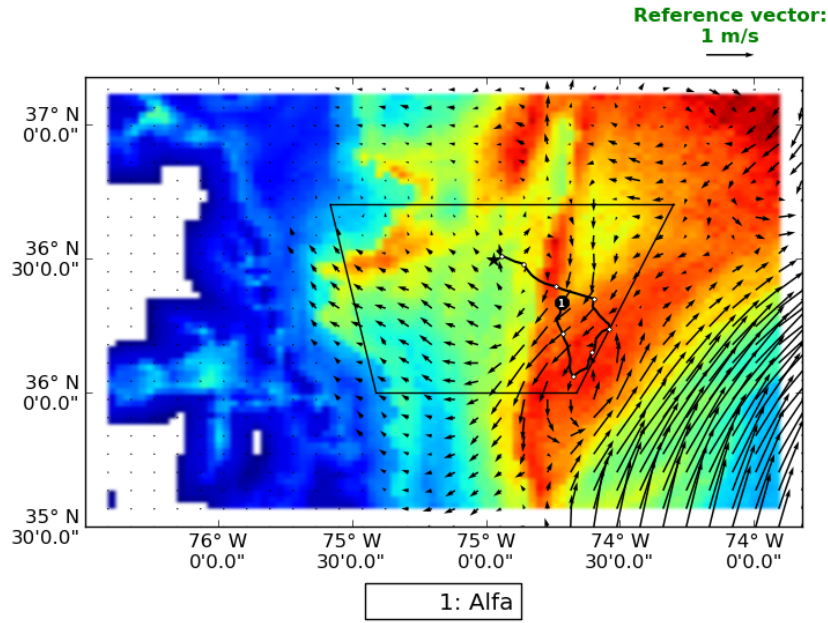

**Figure 4-3:** The suggested glider path (black line with waypoints), rendezvous point (black star), operational area (black trapezoid), water current strength (vector field), and uncertainties (morphology)

## B. Localized Morphology for Single Glider Path

There may be areas of high morphology values that are not reachable for each glider and so it is useful to compare the suggested glider trajectories with local values of the morphology renormalized. Figure 4-4 displays the difference between the original globally-scaled color values versus the newly calculated locally-scaled colors. This allows better evaluations to be made for individual glider paths as it's easier to discern the differences between the possible varying coordinates.

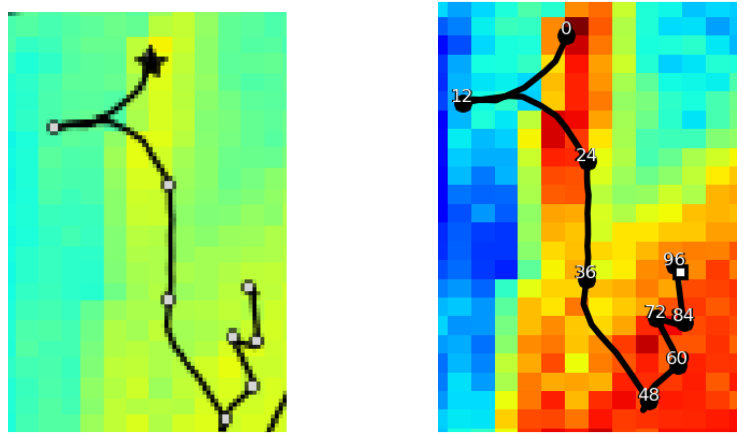

**Figure 4-4: (a)** Suggested path (black line with waypoints), starting position (black star), with global normalization of uncertainties' colors **(b)** Suggested path (black line with waypoints), with localized normalization of uncertainties' colors

## C. Waypoint Files

The waypoint file is the primary deliverable as its purpose is to provide all the necessary information for the gliders' instructions. An example waypoint file for the suggested path in Figure 4-5 is provided in Table 1. A waypoint file is generated by modifying the original EMPath 'Best Run' data *(csv file)*. Each row of the waypoint file contains the data for a waypoint. Waypoints are distinguished by their expected arrival time (in hours) relative to the glider's initial deployment, in Table 4-3, the waypoints occur in 12 hour increments. This file uses *Degrees, Minutes* notation instead of the longitude and latitude decimal notation originally output by EMPath. Lastly, the renormalized numerical values (for the localized morphology) are also provided as reference, as depicted in Table 4-3, giving some insight as to why those waypoints are selected for sampling.

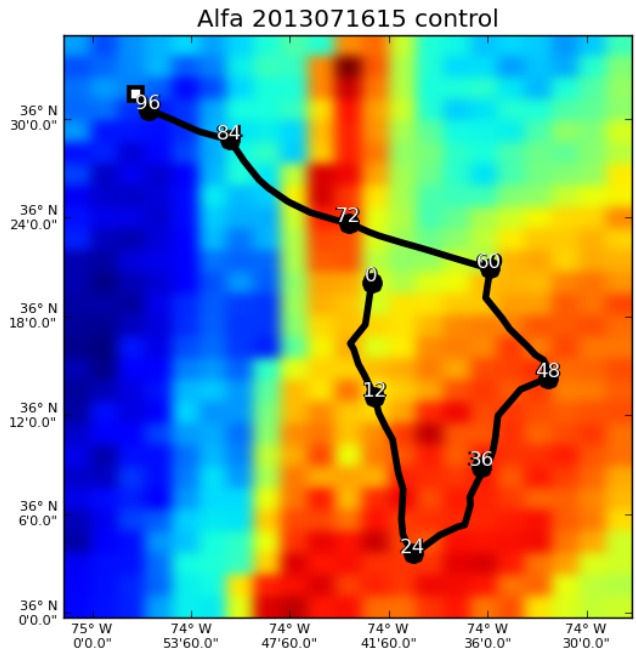

**Figure 4-5:** The graphic to the left has the suggested path (black line with waypoints) and the renormalized uncertainties (morphology).

```
Time, Latitude,      Longitude,       Platform, Out Of Bounds?, Too Close?, Weight
000,  36 N 20'03.840", 74 W 43'01.920", Alfa,     Clear,          Clear,      0.6130
012,  36 N 13'06.960", 74 W 42'50.760", Alfa,     Clear,          Clear,      0.6566
024,  36 N 03'36.000", 74 W 40'29.280", Alfa,     Clear,          Clear,      0.8428
036,  36 N 08'51.000", 74 W 36'21.240", Alfa,     Clear,          Clear,      0.8720
048,  36 N 14'15.720", 74 W 32'20.760", Alfa,     Clear,          Clear,      0.8563
060,  36 N 20'53.880", 74 W 35'52.800", Alfa,     Clear,          Clear,      0.5604
072,  36 N 23'40.560", 74 W 44'25.080", Alfa,     Clear,          Clear,      0.8669
084,  36 N 28'42.600", 74 W 51'38.160", Alfa,     Clear,          Clear,      0.3714
096,  36 N 30'32.760", 74 W 56'33.720", Alfa,     Clear,          Clear,      0.1499
098,  36 N 31'33.600", 74 W 57'21.240", Alfa,     Clear,          Clear,      0.1499
```

**Table 4-3:** The corresponding waypoint file for Figure 4-5; where the last column has the renormalized morphology values

### D. Forecast Animations

An animated sequence, as represented in Figure 4-6, also provides a visual time series analysis expressing changes in water currents and temperature over time. For this animation, the forecasted velocities and temperature field are from the *GMAST_CCF*. The temperature field is useful to evaluate the glider's suggested path with respect to the ocean dynamics and variability. For this research, the uniform time interval per frame is derived from the GMAST_CCF's dynamical function. Since the ocean forecast may be shorter than the time length of the suggested path then the last forecasted frame is used for the remaining frames of the glider animations.

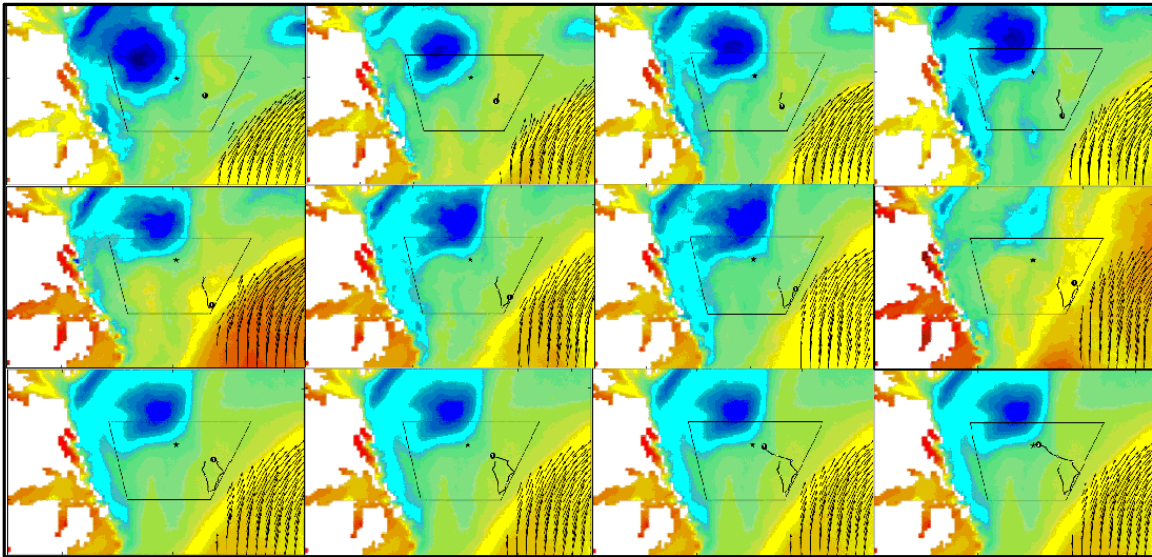

**Figure 4-6:** Suggested glider path (black line with waypoints), rendezvous point (black star), Operational area (black trapezoid), water current strength (vector field), temperatures (colored background), and the time interval between frames is 3 hours.

## 4.2.1 Options

Options are also included for optimizing these visualizations to enhance their reliability in delivering meaningful results. They allow for the imported data to be filtered reducing the effects of either unwanted values or spurious values. They also allow for the imported data to be better fit to align with the graphic's parameter space. The following data can be adjusted to increase the usability of the images:

1. *Crop Boundaries* - the morphology file may have artificial high values at its boundaries as illustrated in Figure 4-7 that may alter the normalized values of the inner domain so an option has been added to crop the boundaries. For consistency, the cropped boundaries are treated as masked values and omitted from the normalization but still drawn as white cells on the visualization. This provides a much greater color gradient for the remaining portion of the morphology, as seen in Figure 4-7.
24

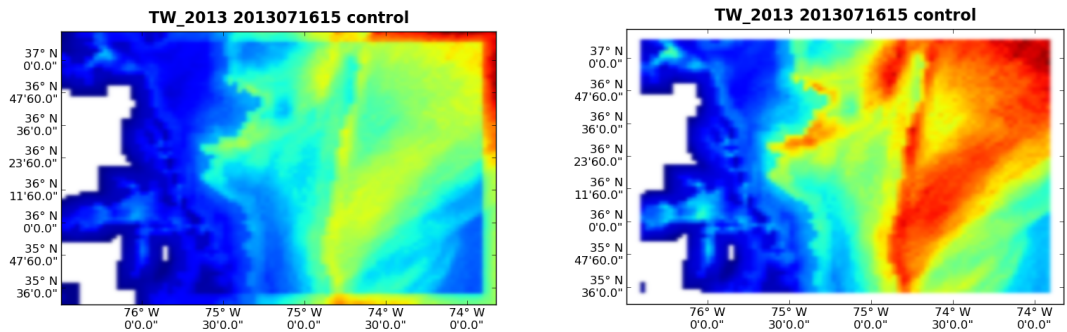

**Figure 4-7: (a)** Original morphology where its boundaries affect the color mapping of the rest of the space **(b)** Cropped morphology masks the boundaries allowing the remaining regions to be colored with greater number of colors

2. *Smooth Image* - the default visualization provides a uniformed coloring for each individual grid point in the morphology. However, the resulting image may have a pixilated look as shown in Figure 4-8a. An option to interpolate is provided, an example is shown in Figure 4-8b. There are 5 different selectable interpolation schemes depending on the level of blending the user desires: None, Kaiser, Bilinear, Gaussian, and Bicubic.

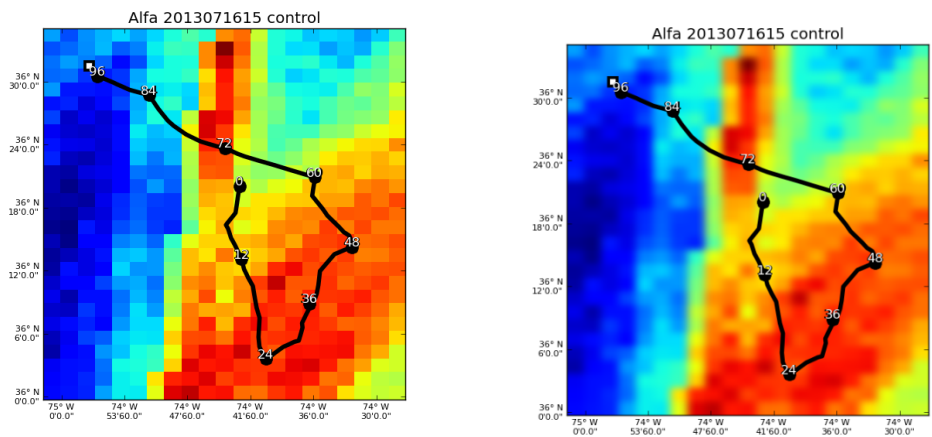

**Figure 4-8: (a)** Path (black line and waypoints) with pixilated morphology where each cell is uniformly colored **(b)** Path (black line and waypoints) with smoothed morphology where each cell is blended with its neighboring color values

3. *Vector Density* - the quantity of 'water current' values provided may be too numerous for the region space to plot them meaningfully. If all vectors were drawn then the vector field could get too cluttered to decipher as shown in Figure 4-9a. So an option to reduce the number of vectors displayed has been provided as shown in Figure 4-9b.

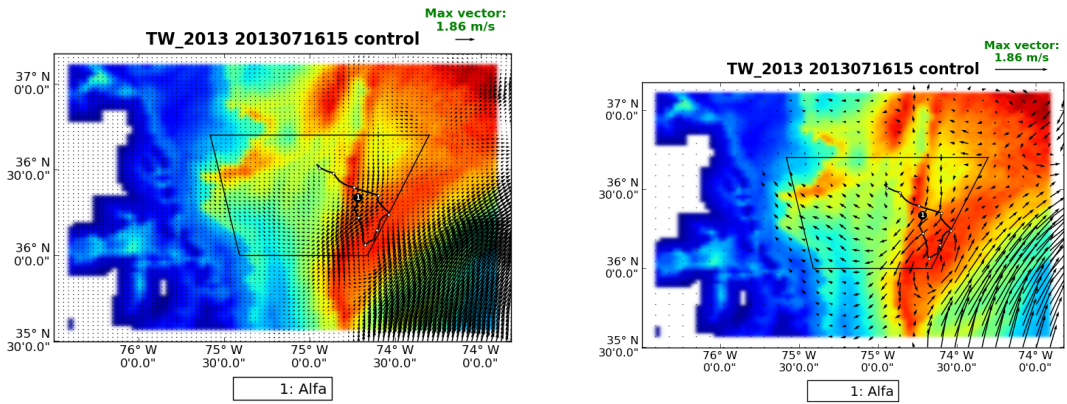

**Figure 4-9: (a)** Path (black line and waypoints), operational area (trapezoid), uncertainties (morphology) with the complete vector field rendered which clutters the image and makes it difficult to understand where as **(b)** has the sparse vector field which draws fewer vectors but becomes much easier to understand

4. *Mask Vectors* - since the vector field is typically only used to locate and analyze the strong currents then there is the option to only draw those currents of a specified magnitude, as shown in Figure 4-10.

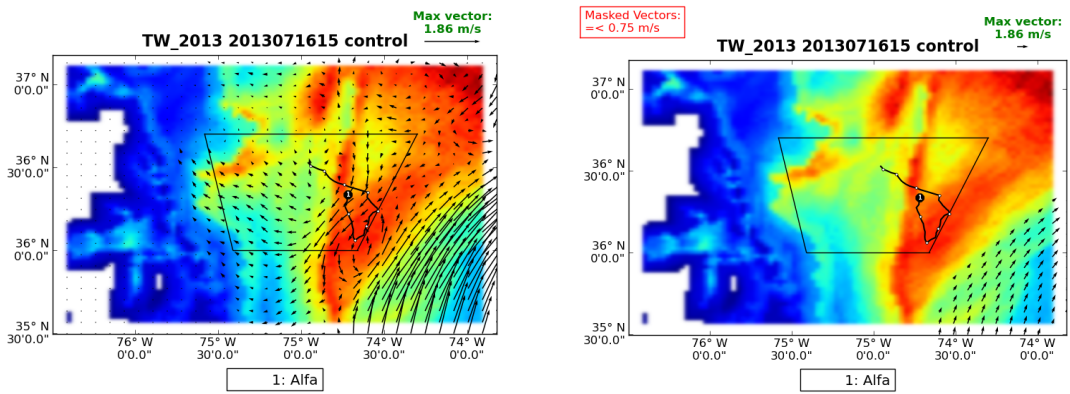

**Figure 4-10: (a)** the non-masked vector field displays vectors of all magnitudes no matter how small. **(b)** The masked vector field only displays the stronger vectors

## 4.3: Evaluate the quality of the waypoints

EMPath executes for multiple times, generating an optimal path for each of its runs henceforth defined as a top run. The best run is the top run with the highest score from which the waypoints are delivered. The objective is to evaluate the best run with respect to the other top runs provided by EMPath. The criteria being evaluated include:
1. Comparing the runs by score
2. Comparing the runs by behavior to determine if they tend to converge

Figure 4-11 shows that there are three possible scenarios dealing with convergence: 1) the paths may converge, 2) the paths diverge into two or multiple paths, or 3) the paths diverge into a spread.

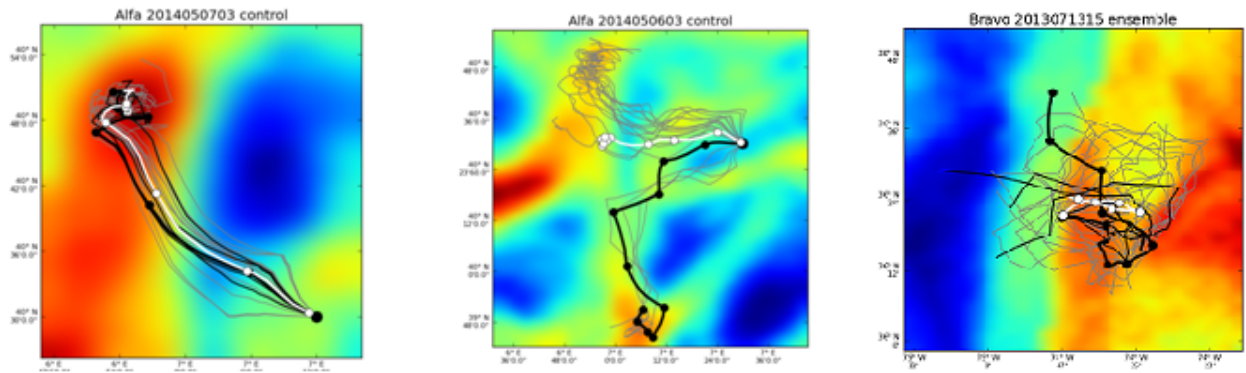

**Figure 4-11: a)** the various suggested paths converge **b)** the various paths diverge into two or more paths. **c)** The various paths diverge into a spread. a, b, and c contains the best path (black), other top paths (grey), mean of all paths (white), and morphology of uncertainties (background colors)

From these observations, the strength of the best run is determined by its spread; the higher the concentration of runs for a waypoint, the greater the confidence is for that waypoint. The area containing the concentration of points is depicted with an ellipse. If the glider resurfaces inside this ellipse then we can speculate that the acquired data would have an impact on the next cycle of data assimilations; otherwise it may not be the case even though the glider is at a close distance from the assigned waypoint. Finally, once the ellipses for all the waypoints are known, they may then be directly compared to one another to see how the confidences change across the best path. The following paragraphs describe the 5 major associated visualizations.

### A. Histogram

A histogram is created to show the distribution of scores to find how comparable or relevant the other runs are in relation to the best run. All of the scores are normalized and then the corresponding runs are subdivided into one of four possible groups:
1. Runs that are higher than 75% of the top score
2. Runs that are between 50% to 75% of the top score
3. Runs that are between 25% to 50% of the top score
4. Runs that are less than 25% of the top score

The histogram is then used to find how many runs are scored similar to the best run, an example shown in Figure 4-12. Those higher scored runs may offer insight into the best path.

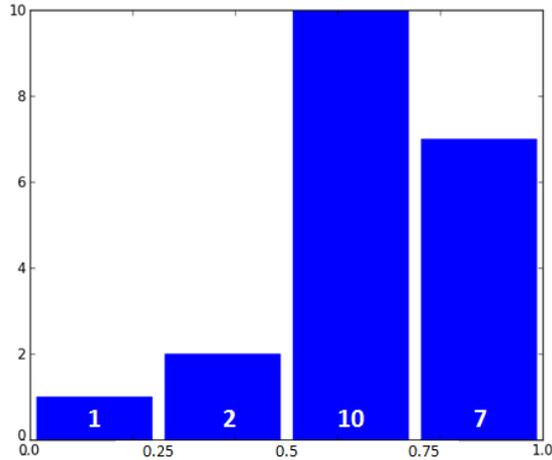
**Figure 4-12:** This histogram shows that one run is in group 4, two runs are in group 3, ten runs are in group 2, and seven runs are in group 1.

### B. Top runs

All the runs generated by EMPath are displayed on top of the morphology (Figure 4-13). Using the histogram from part A, the top 75% of runs are highlighted. If there are more than five runs in the top 75% then only the top five are used. The best run is bolded along with the waypoints. The mean for all runs is also plotted.

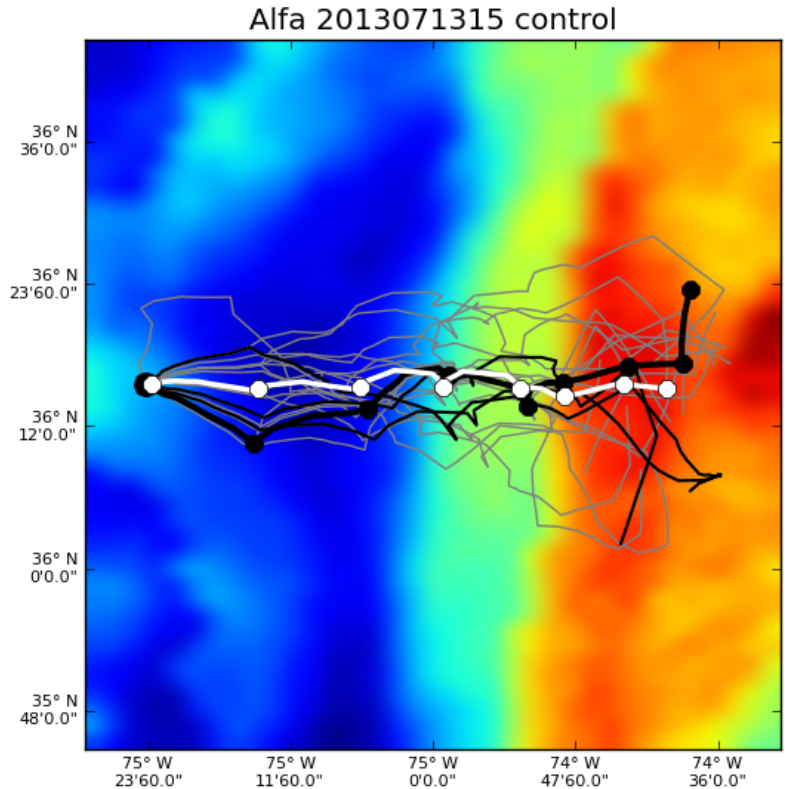

**Figure 4-13:** All top paths (gray line), top 80% (black line), top run (thick black line), waypoints (black dots), mean path (white line), mean at waypoints (white dots) with uncertainties colored (morphology)

## C. Ellipses

Confidence defines the amount of viable space surrounding a waypoint that will still yield valuable data. The goal is to evaluate the spread of the top run points for a measure of the confidence of the EMPath solution and to provide the confidence levels on the quality of the best path. An ellipse is used to visualize the confidence for a waypoint and the method for constructing these confidence ellipses is detailed in chapter 5. After an ellipse is drawn, showing the distribution of the top run points within it helps to understand the ellipse's size, shape and orientation. For each waypoint time, an individual ellipse is constructed (see Figure 4-14). The ellipse's orientation uses the northern axis and a clockwise rotation. The major and minor axes are drawn to show the ellipse's orientation in relation to the northern axes. The center of the ellipse is explicitly shown since all points are plotted relative to it in terms of distance in kilometers. For minimizing the ellipse, isolated points may be ignored, a concept further explained in chapter 5. The points within the ellipse are colored red, green, blue or yellow depending on which quadrant of the ellipse that they appear in. This coloring may offer important insight regarding where the points are concentrated thereby providing a means for subdividing the ellipse, if necessary. Finally, since the ellipse's purpose is to show the confidence of the waypoint, the waypoint (best run position) is also highlighted.

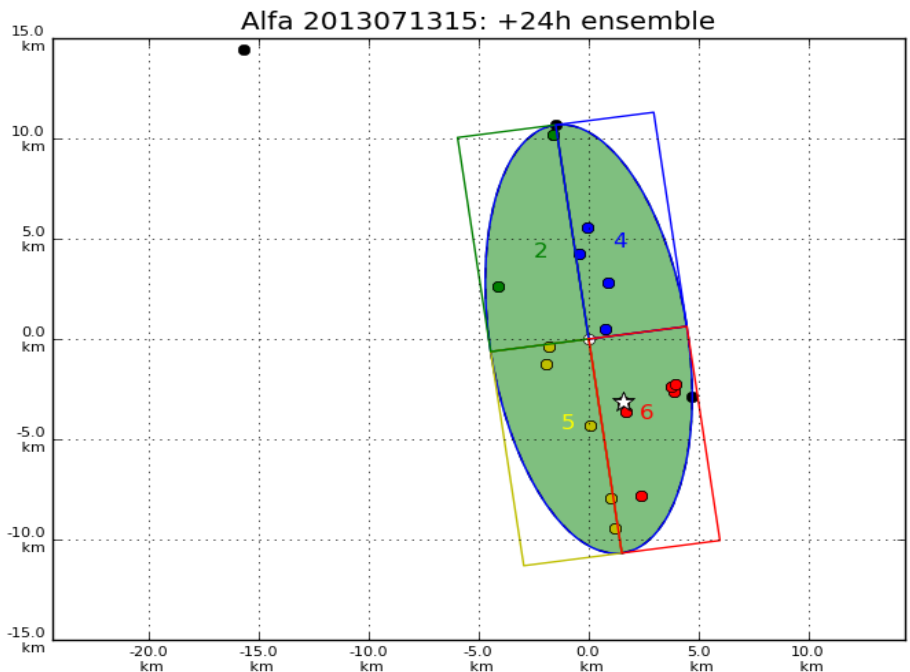

**Figure 4-14:** The ellipse (green), Axes (black lines), center (white dot), ellipse points: quadrant 1 points (yellow dots), quadrant 2 points (green dots), quadrant 3 points (blue dots), quadrant 4 points (red dots), outliers (black dots)

## D. All Ellipses Map

Figure 4-15 shows the confidence for all the waypoints relative to one another. This way the size of each ellipse can be directly compared. The ellipses where the points tend to converge are smaller in size and the ellipses where the points greatly vary are larger in size. This is accomplished by having all of the ellipses plotted along with the suggested path on top of the localized morphology. The delivered waypoints are highlighted as well as the center points for each ellipse.

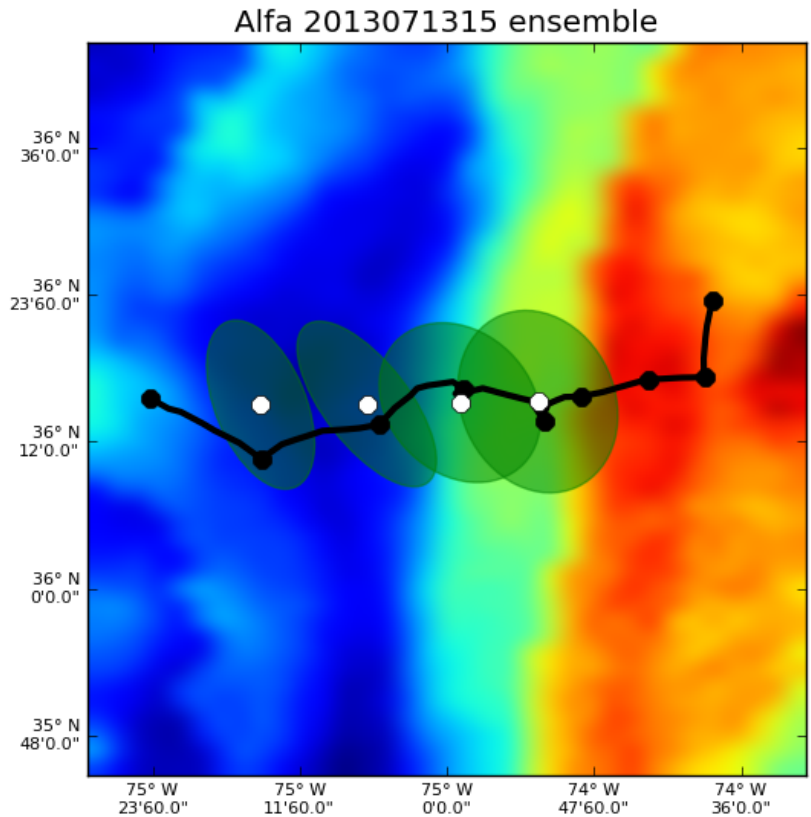

**Figure 4-15:** Top path (black line), waypoints (black dots), confidence ellipses (green), ellipse center points (white dots) with the uncertainties colored (morphology)

### E. Confidence file

The confidence file, (Table 4-5), contains the numerical data for which all the ellipses are defined. The purpose is to allow the underlining values used to produce these deliverables be readily available and reconstructable for any further analysis. The confidence file is written as a common text file using a comma-separated value format. It contains each glider's name and the corresponding waypoint times, the center points' longitude and latitude values, the major radii in kilometers, minor radii in kilometers, and the angle. Angle is defined for the major radius, in degrees, using north as the origin axis, moving in a clock-wise rotation.

```
Platform, Time, Latitude, Longitude,     DR,      dr,   Angle
   Alfa,   12, 36 N 15', 75 W 15',  8.99km,  1.80km,   2.171
   Alfa,   24, 36 N 15', 75 W 06', 10.78km,  4.50km, 171.974
   Alfa,   36, 36 N 15', 74 W 58',  8.83km,  6.67km, 157.570
   Alfa,   48, 36 N 15', 74 W 52', 10.77km,  5.75km, 157.773
  Bravo,   12, 36 N 23', 74 W 43', 14.58km,  9.60km, 151.796
  Bravo,   24, 36 N 23', 74 W 39', 24.29km, 18.17km, 165.115
  Bravo,   36, 36 N 22', 74 W 35', 26.86km, 17.55km,  13.928
  Bravo,   48, 36 N 21', 74 W 35', 27.98km, 25.76km,  49.266
  Clint,   12, 36 N 35', 74 W 28',  9.10km,  5.16km, 159.319
  Clint,   24, 36 N 31', 74 W 36', 12.21km,  6.32km, 158.943
  Clint,   36, 36 N 25', 74 W 40', 16.82km, 13.52km, 142.708
  Clint,   48, 36 N 21', 74 W 40', 18.48km, 15.40km, 146.811
```

**Table 4-5:** The confidence file with glider name, time, center point, major and minor radii, and angle

### 4.3.1 Options

1. *Start Hour* - this parameter is used to select a starting time for which to generate the evaluations. The expected integer value is in terms of hours after the last set of glider instructions. Only those waypoints that occur after the start hour are used for the deliverables. The default value is set to 12 hours.

2. *Stop hour* - this parameter is used to select a stopping time for which to generate the evaluations. The expected integer value is in terms of hours after the last set of glider instructions. Only those waypoints that occur before the stop hour are used for the deliverables. The default value is set to 48 hours.

3. *STD* - short for standard deviations, is the parameter to adjust the confidence interval which determines the outlier points. A value of 0 represents excluding all points except for the center point. The higher the number the more points it will include. A value of 3 should include all points. This parameter is also referred to as the *scale factor*. The default value for STD is 1.5.

# 5 Building a minimal-fit confidence ellipse

## 5.1 Approach

There are only 3 necessary components required for constructing an ellipse: center point, major axis, and minor axis. The angle of ellipse is implicitly given by the orientations. Constructing a minimal-fit ellipse of the top runs is no trivial task. There are many considerations that should be made that may affect the resulting ellipse's shape, size, and orientation.

**1. Choice of coordinate system**
First the underlining space or coordinate system used for building an ellipse must be defined. EMPath provides points in terms of longitude and latitude values, (i.e. a spherical coordinate system). Since the longitude and latitude axes are not equally scaled, a direct mapping into a 2d spherical space would not accurately reflect the physical distance. For this reason, the ellipses are plotted in a Cartesian coordinate system and the values expressed in kilometers. Figure 5-1 compares the ellipses in the two different coordinate systems.

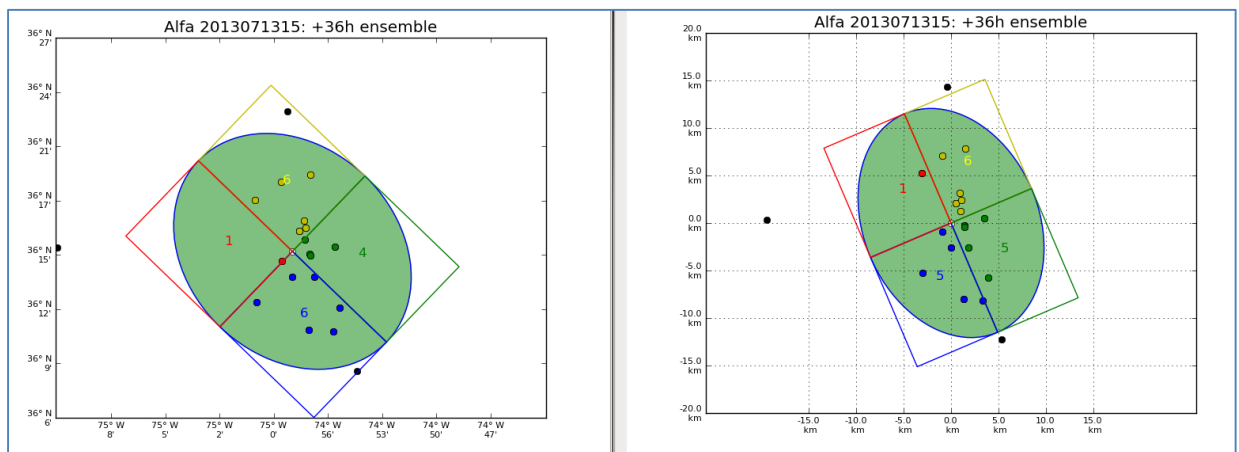

**Figure 5-1 a)** Ellipse plotted with the lon and lat cords; **b)** Ellipse plotted using kilometers

**2. Center point**
The goal is to find the minimal-fit for all points; therefore the natural choice is the use of the mean of all the top run points rather than the waypoint (best run's position). This is because the mean represents the central tendency, which is an important attribute in keeping the ellipse minimal. In fact, it is possible for the waypoint to actually be an isolated point with regard to the other top solutions, as shown in Figure 4-11b.

**3. Defining Outliers**
Outliers are points that are well outside of the expected range of values. In this research, outlier values are henceforth defined in terms of standard deviations (STD) steps away from the mean (center). Any point that falls outside of the stipulated STD threshold (i.e. user parameter) is considered an outlier and may be discarded from the data set.

## 4. Defining the axes
Two basic methods were used for finding the ellipse's axes.

### A. Covariance ellipse
This method uses a covariance matrix and the eigenvalues with the corresponding eigenvectors to ascertain the ellipse's attributes (Hoover, 1984). A more detailed explanation is in appendix A.1. An example ellipse constructed from this method is shown in Figure 5-2a.

The covariance matrix is constructed from the standard deviation values. A linear transformation on the matrix yields eigenvalues and the corresponding eigenvectors. The larger eigenvalue is the major radius length and the smaller eigenvalue is the minor radius length. The orientation of the ellipse is calculated using the eigenvector that belongs to the major radius. The eigenvector provides the point's values required to rotate the axes into position.

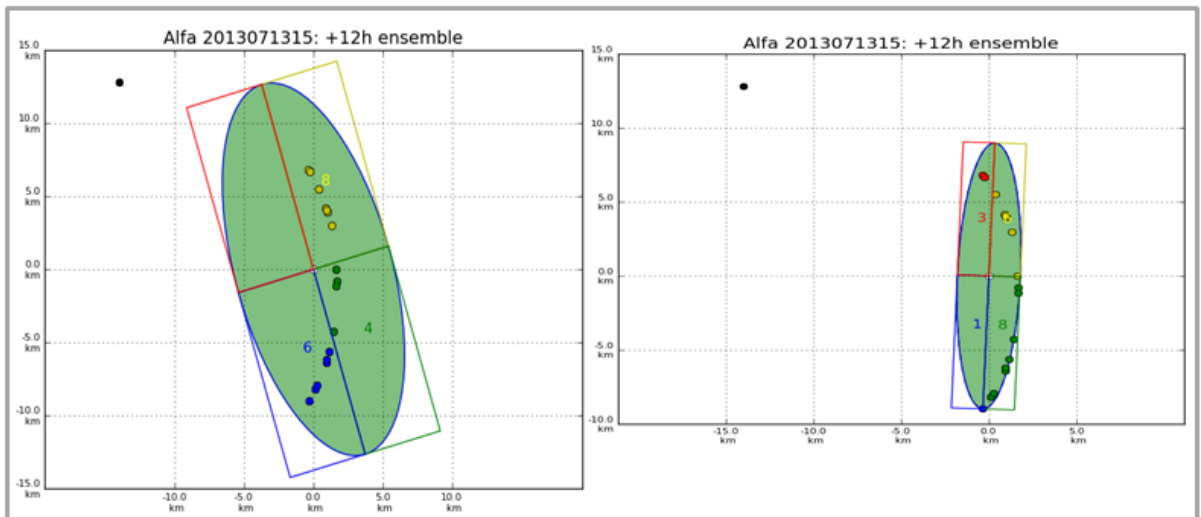

**Figure 5-2: a)** Covariance Ellipse - the origin point (white) and the ellipse (green) are fit around a set of points (yellow, green, blue). The omitted outlier is black. **b)** Point-fitting Ellipse - the origin point (white) and the ellipse (green) are fit around a set of points (yellow, green, blue). The omitted outlier is black.

### B. Point-fitting ellipse
This method removes the outlier points and then uses an algorithmic approach to fit the ellipse directly to the remaining points using the Euclidean distance. A more rigorous explanation for this approach is provided in Appendix A.2. An example ellipse constructed using this method is shown in Figure 5-2b.

Given a cluster of points, the major radius length is calculated as the distance from the origin point (mean) and the furthest point. The minor radius is found using an iterative process. First, the worst-case scenario is initialized whereby the minor radius is equal to the major radius (i.e. a circle). Then the minor radius is continually contracted in until it cannot get any smaller without omitting non-outlier points. The final contracted value when this end condition occurs is the minor radius length.

## 5.2 Comparison between methods A and B

Method B may outperform method A in instances similar to Figure 5-3, where there is an extreme outlier, a single point that is a great distance from the point cluster. That single outlier impacts the covariance to such a degree as to enlarge the ellipse and alter its orientation. Whereas, the point-fitting method ignores that outlier point altogether, calculating the angle and size of the ellipse independent of it. In Table 5-1, the confidence tables for method A and method B containing the numerical values for the ellipse radii are provided.

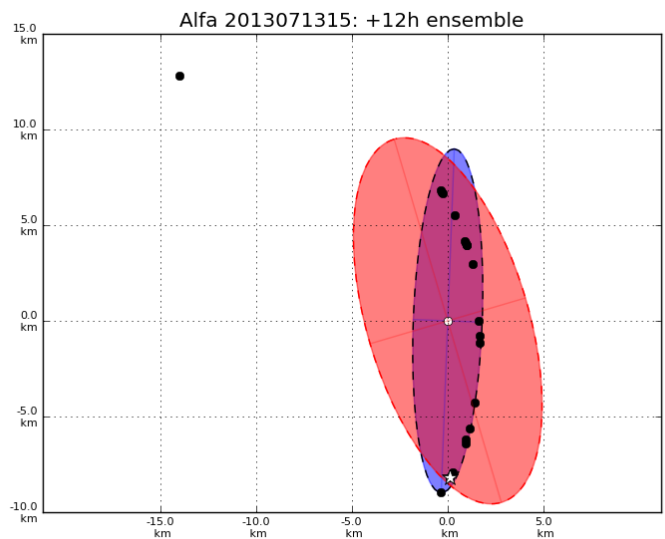

**Figure 5-3:** The method 1 ellipse (red ellipse) and method 2 (blue ellipse) are plotted around the best runs (black dots); these ellipses are for the 12th hour waypoint.

### Confidence Table Method A

```
Platform, Time, Latitude,  Longitude, DR,      dr,      Angle
Alfa,     12,   36 N 15',  75 W 15',  9.87km,  4.29km,  -22.411
Alfa,     24,   36 N 15',  75 W 06',  10.50km, 4.24km,  -37.802
Alfa,     36,   36 N 15',  74 W 58',  9.07km,  7.32km,  -46.063
Alfa,     48,   36 N 15',  74 W 52',  10.32km, 7.32km,  -24.473
Bravo,    12,   36 N 23',  74 W 43',  13.84km, 10.18km, -42.024
Bravo,    24,   36 N 23',  74 W 39',  24.84km, 11.21km, -11.62
Bravo,    36,   36 N 22',  74 W 35',  26.41km, 10.67km,  13.241
Bravo,    48,   36 N 21',  74 W 35',  26.39km, 11.73km,  43.414
Clint,    12,   36 N 35',  74 W 28',  9.24km,  2.96km,  -13.799
Clint,    24,   36 N 31',  74 W 36',  13.24km, 4.23km,  -35.776
Clint,    36,   36 N 25',  74 W 40',  17.27km, 10.59km, -57.50
Clint,    48,   36 N 21',  74 W 40',  19.97km, 11.26km, -60.808
```

### Confidence Table Method B

```
Platform, Time, Latitude,  Longitude, DR,      dr,      Angle
Alfa,     12,   36 N 15',  75 W 15',  8.99km,  1.80km,   2.171
Alfa,     24,   36 N 15',  75 W 06',  10.78km, 4.50km,  171.974
Alfa,     36,   36 N 15',  74 W 58',  8.83km,  6.67km,  157.570
Alfa,     48,   36 N 15',  74 W 52',  10.77km, 5.75km,  157.773
Bravo,    12,   36 N 23',  74 W 43',  14.58km, 9.60km,  151.796
Bravo,    24,   36 N 23',  74 W 39',  24.29km, 18.17km, 165.115
Bravo,    36,   36 N 22',  74 W 35',  26.86km, 17.55km,  13.928
Bravo,    48,   36 N 21',  74 W 35',  27.98km, 25.76km,  49.266
Clint,    12,   36 N 35',  74 W 28',  9.10km,  5.16km,  159.319
Clint,    24,   36 N 31',  74 W 36',  12.21km, 6.32km,  158.943
Clint,    36,   36 N 25',  74 W 40',  16.82km, 13.52km, 142.708
Clint,    48,   36 N 21',  74 W 40',  18.48km, 15.40km, 146.811
```

**Table 5-1:** A.) Confidence Table for method A, highlighting the row for glider: Alfa and hour 12
B.) Confidence Table for method B, highlighting the row for glider: Alfa and hour 12

On the other hand, method A may outperform method B, as shown in Figure 5-4. This occurs because method B first removes the outlier points using an assumed orientation of 0. However, if the resulting ellipse's orientation differs, then there may be points that were included which should have been removed resulting in an ellipse that is larger when compared to method A. In Table 5-2, the confidence tables for method A and method B containing the numerical values for the ellipse radii are provided.

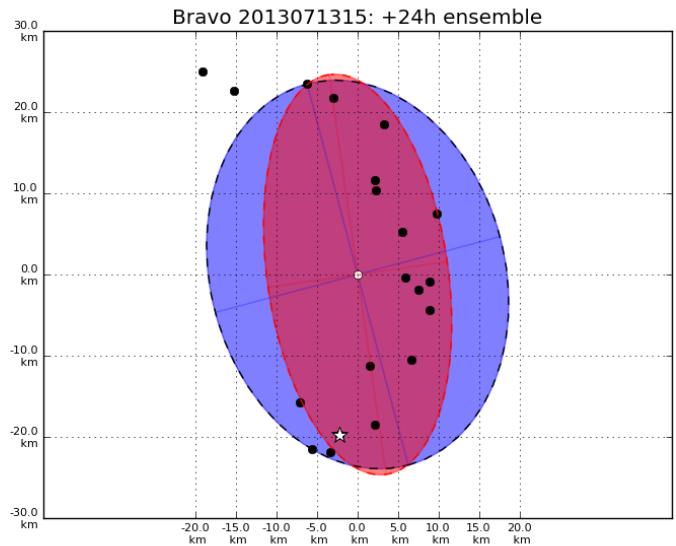

**Figure 5-4:** The method 1 ellipse (red ellipse) and method 2 (blue ellipse) are plotted around the best runs (black dots); these ellipses are for the 24th hour waypoint

### Confidence Table Method A

```
Platform, Time, Latitude, Longitude, DR,      dr,     Angle
Alfa,     12,   36 N 15', 75 W 15',   9.87km,  4.29km, -22.411
Alfa,     24,   36 N 15', 75 W 06',  10.50km,  4.24km, -37.802
Alfa,     36,   36 N 15', 74 W 58',   9.07km,  7.32km, -46.063
Alfa,     48,   36 N 15', 74 W 52',  10.32km,  7.32km, -24.473
Bravo,    12,   36 N 23', 74 W 43',  13.84km, 10.18km, -42.024
Bravo,    24,   36 N 23', 74 W 39',  24.84km, 11.21km, -11.62
Bravo,    36,   36 N 22', 74 W 35',  26.41km, 10.67km,  13.241
Bravo,    48,   36 N 21', 74 W 35',  26.39km, 11.73km,  43.414
Clint,    12,   36 N 35', 74 W 28',   9.24km,  2.96km, -13.799
Clint,    24,   36 N 31', 74 W 36',  13.24km,  4.23km, -35.776
Clint,    36,   36 N 25', 74 W 40',  17.27km, 10.59km, -57.50
Clint,    48,   36 N 21', 74 W 40',  19.97km, 11.26km, -60.808
```

### Confidence Table Method B

```
Platform, Time, Latitude, Longitude, DR,      dr,     Angle
Alfa,     12,   36 N 15', 75 W 15',   8.99km,  1.80km,   2.171
Alfa,     24,   36 N 15', 75 W 06',  10.78km,  4.50km, 171.974
Alfa,     36,   36 N 15', 74 W 58',   8.83km,  6.67km, 157.570
Alfa,     48,   36 N 15', 74 W 52',  10.77km,  5.75km, 157.773
Bravo,    12,   36 N 23', 74 W 43',  14.58km,  9.60km, 151.796
Bravo,    24,   36 N 23', 74 W 39',  24.29km, 18.17km, 165.115
Bravo,    36,   36 N 22', 74 W 35',  26.86km, 17.55km,  13.928
Bravo,    48,   36 N 21', 74 W 35',  27.98km, 25.76km,  49.266
Clint,    12,   36 N 35', 74 W 28',   9.10km,  5.16km, 159.319
Clint,    24,   36 N 31', 74 W 36',  12.21km,  6.32km, 158.943
Clint,    36,   36 N 25', 74 W 40',  16.82km, 13.52km, 142.708
Clint,    48,   36 N 21', 74 W 40',  18.48km, 15.40km, 146.811
```

**Table 5-2:** A) Confidence Table for method A, highlighting the row for glider: Bravo and hour 24
B) Confidence Table for method B, highlighting the row for glider: Bravo and hour 24

## 5.3 Deliverable: a combination of methods A and B

For this final method, only those points contained within the method A ellipse are used to perform the point-plotting fit of method B. This does an even better job of finding a minimally sized ellipse and increases the confidence level around the mean and the waypoint. Figure 5-5a shows an example of where method A outperforms method B, but then method A+B outperforms method A in Figure 5-5b.

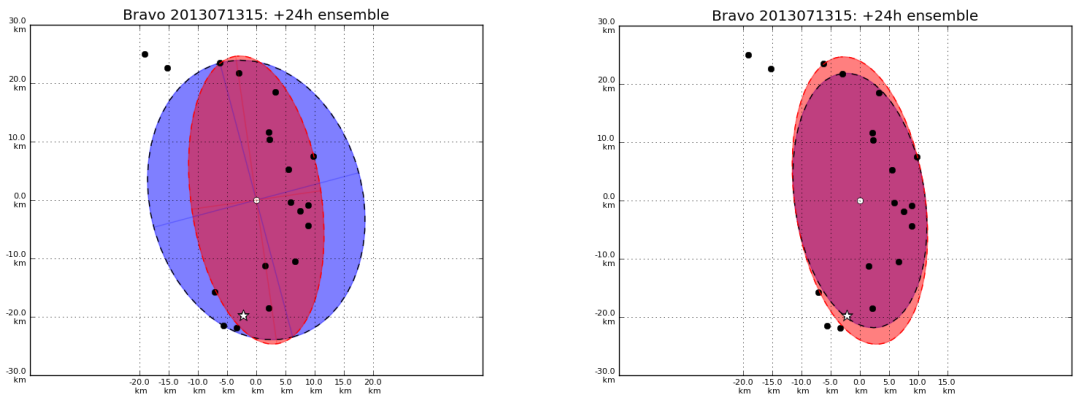

**Figure 5-5 a)** The method 1 ellipse (red ellipse) and method 2 (blue ellipse) are plotted around the best runs (black dots) **b)** the method 1 ellipse (red ellipse) and method 3 (blue ellipse) are plotted around the best runs (black dots); these ellipses are for the 24th hour waypoint

## 5.4 The sensitivity of the ellipse to STD values

The ellipse size and shape is sensitive to outlier points so that it can be altered based upon how many standard deviations away from the center are included. The sensitivity is adjusted by either increasing or decreasing the number of STD steps that should be included. The underlining mathematical formulas calculate for a 1σ ellipse (1 STD). However, the method is generalized to include a predefined STD values. This is accomplished by multiplying the major and minor axes with a scale factor. Figure 5-6 illustrates a 1σ, 1.5σ, and 2σ ellipse respectively, since the original scale is 1 STD, the new scale factor need only be 2 for a 2σ ellipse or 1.5 for a 1.5σ ellipse. The default scale factor is 1.5, which tends to includes most points removing only the outliers.

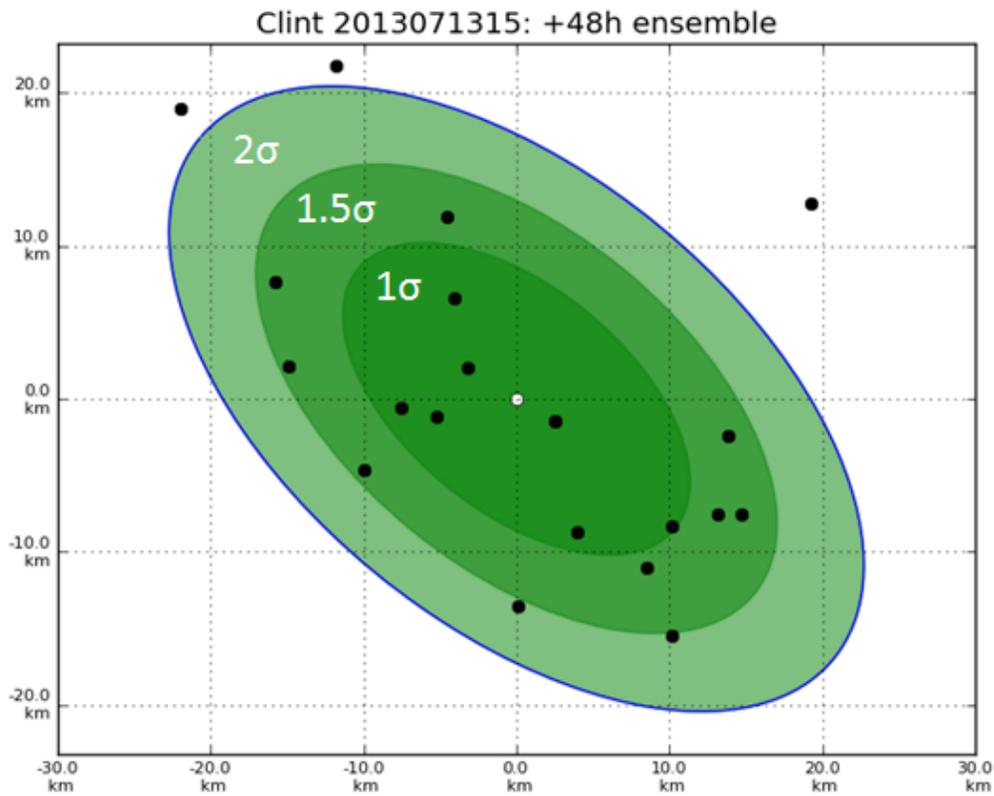

**Figure 5-6** The origin point (white) and the set of points (black), the smallest ellipse is a 1σ ellipse, the middle ellipse is a 1.5σ ellipse, and the largest ellipse is a 2σ ellipse.

# 6. Summary and Discussions

The motivation for this research is to contribute towards improving the forecasting skills for the NRL's RTOFS called RELO. RELO forecasts future ocean states by a combination of oceanographic observations along with a background field. NCODA, a data assimilation scheme, merges the data and background field together and applies forcing to derive an initial state from which NCOM, a dynamical model, predicts a future. However, uncertainties within the model may limit the prediction's accuracy. Adaptive sampling is one way to improve RELO; the approach is to target the next set of observations into areas where the forecast field yields the most uncertainties. Adaptive sampling is mostly implemented with underwater gliders. A GA is applied to find a near-optimal path to maximize the use of the gliders. The GA's suggested path is provided as a set of waypoints or coordinates and is meant only to provide guidance. The problem is that there is no unified, robust set of tools by which to evaluate the GA's output to better enable the adaptive sampling. Currently, there is a need to bridge the gulf from the modeling to the operative environments.

The objective is to better enable the operative team in performing adaptive sampling by providing a robust visualization toolset used to evaluate the GA's suggested path against any additional operative criteria. Visualization is required since humans must manually perform these evaluations. Visualizations allow for a quicker and easier means toward understanding the data. The visual evaluation toolset is subdivided into three packages that may be used at different times. This allows for each package to focus on a particular goal, providing only that data necessary to support a specific phase in the adaptive sampling strategy rather than inundating the user with all the data at once. The first package is to verify that the glider is actually following the waypoints and to predict the position of the glider for the next cycle's instructions. The second package helps ensure that the delivered waypoints are both useful and feasible. The third package provides the confidence levels for the suggested path. These visualization packages are written in the Python programming language because it is a general-purpose language that is also suitable for scientific computing. The software is designed to have an easy interface for user input. An additional aim for this software package is to provide the capability for easy expansion, i.e. adding new and different future visualizations. This is realized through a modular programming approach that builds base modules for each of the RELO data components (i.e. morphology, glider paths, water currents, etc.) in order to access them within the toolset so that the final visualizations are created by simply combining these base modules together.

One of the major challenges for this software is that it has to be integrated into the existing operational system. This necessitated a design that could interface without disrupting and taking advantage of any pre-existing configurations. Another major requirement for this software is portability: i.e. deploy and execute on a wide range of computing platforms and OS. This requirement has already been partially satisfied with early versions that have been delivered to NRL and has been transitioned to NAVO in support of real-time glider exercises. Due to time restrictions this research could not be fully tested on a real-time exercise, although, it has been conducted on a limited scale. A one-way exercise was performed whereby instructions are initially provided, but no follow-up updates occurred.

Future goals aim to improve both the physics and visuals for this toolset. The physics may be improved by incorporating the predicted ocean velocities instead of assuming that the ocean is at rest; this could increase the accuracy of predicted glider movements and other dynamical features. The visuals may also be improved by offering more than just png images. In the next version, these visuals will also be plotted on Google Maps. Additional options for specifying different waypoint definitions are also planned; currently waypoints are strictly time oriented, but in future releases, waypoints may also be identified as those points in the path where major turns occur.

Future versions will also better integrate these set of packages into the operational system. While testing, it became apparent that a smoother transition from the parameters of GMAST and EMPath is needed. This software needs to consistently access and stream new sets of data without the overhead of requiring any manual updates from the user. A solution for interfacing with the system to update the parameters has already been partially tested and developed.

# APPENDIX A: Ellipses fitting

## A.1 Method A: Covariance ellipses

Given $N$ top runs, let $x_i$ and $y_i$, be the coordinates of the top run points at a given time, where $i = 1 \ldots N$. Let $\mu_x, \mu_y$ be their mean, respectively. Then let the covariance, $\sigma_{xy}$, and the variances, $\sigma_{xx}$ and $\sigma_{yy}$, be defined as:

$$\sigma_{xy} = \frac{1}{N} \sum_{i=1}^{N} (x_i - \mu_x)(y_i - \mu_y)$$

$$\sigma_{xx} = \frac{1}{N} \sum_{i=1}^{N} (x_i - \mu_x)(x_i - \mu_x) = \frac{1}{N} \sum_{i=1}^{N} (x_i - \mu_x)^2 = (\sigma_x)^2$$

$$\sigma_{yy} = \frac{1}{N} \sum_{i=1}^{N} (y_i - \mu_y)(y_i - \mu_y) = \frac{1}{N} \sum_{i=1}^{N} (y_i - \mu_y)^2 = (\sigma_y)^2$$

The variance describes the amount of spread or dispersion of the quantity around its own mean value (Vermeer, 2014) and the covariance is a measure of how much the two variables ($x_i$, $y_i$) change together. The variance and covariance values are then used to construct the covariance matrix, $covmat$.

$$covmat = \begin{vmatrix} Var\{x\} & Cov\{x,y\} \\ Cov\{x,y\} & Var\{y\} \end{vmatrix} = \begin{vmatrix} (\sigma_x)^2 & \sigma_{xy} \\ \sigma_{xy} & (\sigma_y)^2 \end{vmatrix}$$

Finally, the eigenvalues $\lambda_1$ and $\lambda_2$ and eigenvectors of the covariance matrix define the ellipse's axes (Smith, 2002); such as:

$$major\ axis = 2\sqrt{\max(\lambda_1, \lambda_2)} \cdot scalefactor$$
$$minor\ axis = 2\sqrt{\min(\lambda_1, \lambda_2)} \cdot scalefactor$$

Where the scale factor is the STD threshold and the corresponding eigenvectors directions is the orientation of the ellipse.

## A.2 Method B: Point-fitting ellipse

Let the center of the ellipse; $\mu_x, \mu_y$ be defined as the mean of all the top run positions.

This approach is to construct an ellipse containing only those points that are within the predefined $scalefactor$ value. This method has two stages: A) finding the major axis and B) finding the minor axis.

### A) Finding the major axis

For the major radius use the furthest point from the origin point, as shown in Figure A-1, where the term *furthest* is defined in Euclidean distance which requires for the differences in x and y values to be independently calculated.

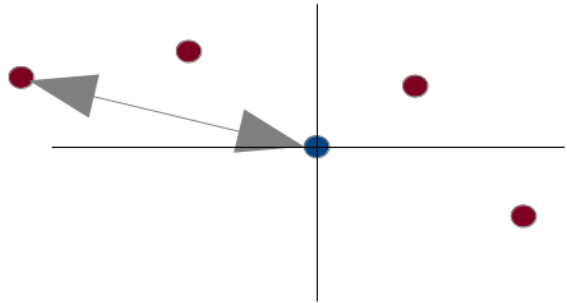

**Figure A-1:** The origin point (blue) and the set of points (red), the major radius (arrow) is drawn between the origin and the furthest point

To find the furthest distance, let $\vec{X}$ and $\vec{Y}$ be defined as:
$$\vec{X} = \{(x_1 - \mu_x), (x_2 - \mu_x), (x_3 - \mu_x), \ldots, (x_n - \mu_x)\}$$
$$\vec{Y} = \{(y_1 - \mu_y), (y_2 - \mu_y), (y_3 - \mu_y), \ldots, (y_n - \mu_y)\}$$

The distance formula is then applied on the two vectors, $\vec{X}$ and $\vec{Y}$ and the maximum resultant is the length of the major radius.

$$major\ radius = \max\left(\sqrt{\vec{X}^2 + \vec{Y}^2}\right)$$

Then the major axis is:
$$major\ axis = 2\ (major\ radius)$$

### B) Finding the minor axis

Finding the minor radius is done algorithmically. It is a 3 step process.

Step 1: Initialize the ellipse as a circle where the minor radius is equal to the major radius, as shown in Figure A-2.

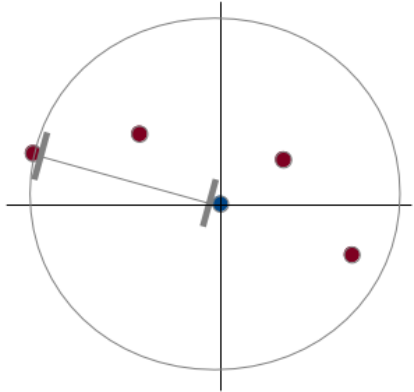

**Figure A-2** The origin point (blue) and the set of points (red), the major radius (line) is drawn between the origin and the furthest point, the initial ellipse (circle) is drawn around the origin

Step 2: Contract the circle in the directions perpendicular to the major axis. This minor radius is decremented by a small constant value $c$, which represents the smallest, finite change between the current ellipse and the next attempted ellipse, see Figure A-3. For this research c= 0.005

$$new\ minor\ radius = current\ minor\ radius - c$$

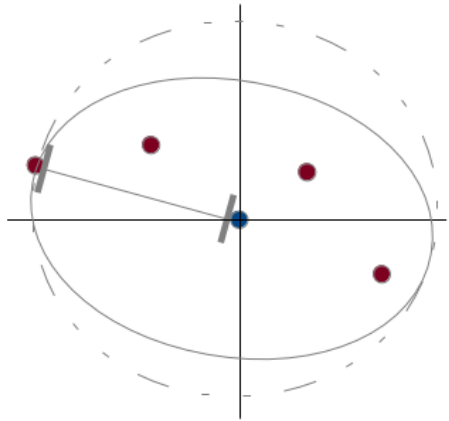

**Figure A-3** The origin point (blue) and the set of points (red), the major radius (line) is drawn between the origin and the furthest point, the initial ellipse (dotted circle), the new ellipse (solid circle)

Step 3: Check that all data points are enclosed within the smaller ellipse. For optimization, the ellipse formula is performed on the vectors $\vec{X}$ and $\vec{Y}$. If any resultant value from this operation exceeds 1 then a point has fallen outside of the ellipse. Otherwise repeat step 2.

$$result = \frac{\left(\left(\vec{X}\cdot\cos\theta\right) + \left(\vec{Y}\cdot\sin\theta\right)\right)^2}{(major\ radius)^2} + \frac{\left(\left(\vec{X}\cdot\sin\theta\right) - \left(\vec{Y}\cdot\cos\theta\right)\right)^2}{(minor\ radius)^2} \begin{cases} goto\ step\ 2; & if\ \max(result) \leq 1 \\ stop\ loop; & if\ \max(result) > 1 \end{cases}$$

An example of the final ellipse using this iterative process is displayed in Figure A-4.

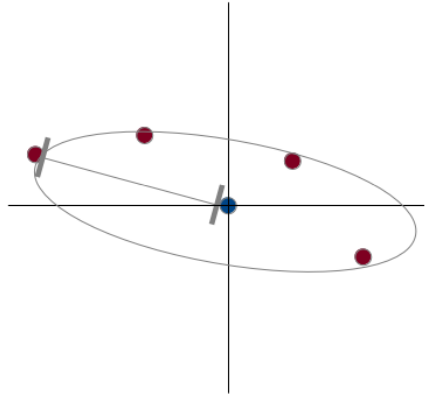

**Figure A-4** The origin point (blue) and the set of points (red), the major radius (line) is drawn between the origin and the furthest point, the final ellipse (solid circle)

Then the minor axis is:
$$minor\ axis = 2 \cdot minor\ radius$$

# Vita

The author was born in New Orleans, Louisiana. He obtained his Bachelor's degree in 2006 from the University of New Orleans. He joined the University of New Orleans computer science graduate program in 2012 and worked as a research assistant under Dr. Germana Peggion of the UNO physics department, working on the EMPath visualization project as part of his computer science thesis.